\documentclass{article} 
\usepackage{iclr2026_conference,times}

\usepackage[utf8]{inputenc} 
\usepackage[T1]{fontenc}    
\usepackage{upquote}  
\usepackage{upgreek}  

\newcommand{\capfont}{\small} 

\newcommand{\dataname}{LVP-1M }
\usepackage[pagebackref=true,breaklinks=true,colorlinks,citecolor=gray]{hyperref}
\usepackage{amsfonts}       
\usepackage{nicefrac}       
\usepackage{microtype}      
\usepackage{titlesec}

\usepackage{amsmath}
\usepackage{amssymb}
\usepackage{mathtools}
\usepackage{amsthm}
\usepackage{booktabs} 
\usepackage{multirow} 
\usepackage{graphicx}
\usepackage{wrapfig}  
\usepackage{tabularx}
\usepackage[table]{xcolor}
\usepackage{amssymb}
\usepackage{setspace}
\usepackage{enumitem}
\usepackage[font={small}]{caption}
\usepackage{wrapfig}
\usepackage{scalerel}
\usepackage{tabu}
\usepackage{titlesec}
\usepackage{float}
\usepackage{makecell}
\usepackage{url}

\renewcommand{\paragraph}[1]{\noindent\textbf{#1}.}

\titlespacing\section{0pt}{3pt plus 1pt minus 1pt}{2pt plus 1pt minus 1pt}
\titlespacing\subsection{0pt}{3pt plus 1pt minus 1pt}{2pt plus 1pt minus 1pt}

\usepackage{algorithm}
\usepackage{algpseudocode}
\usepackage[nottoc]{tocbibind}
\usepackage{minitoc}

\usepackage{graphicx} 

\title{Large Video Planner Enables Generalizable Robot Control}
\author{
Boyuan Chen$^{1*}$ \quad
Tianyuan Zhang$^{1*}$ \quad
Haoran Geng$^{2*}$ \quad
Caiyi Zhang$^{2*}$ \quad
Peihao Li$^{2*}$ \\
\textbf{Kiwhan Song$^{1}$} \quad
\textbf{William T. Freeman$^{1}$} \quad
\textbf{Jitendra Malik$^{2}$} \quad
\textbf{Pieter Abbeel$^{2}$} \quad
\textbf{Russ Tedrake$^{1}$} \\
\textbf{Vincent Sitzmann$^{1}$} \quad
\textbf{Yilun Du$^{3}$} \\
$^{1}$MIT \quad\quad
$^{2}$UC Berkeley \quad\quad
$^{3}$Harvard \\
}

\iclrfinalcopy
\begin{document}

\maketitle

\begin{abstract}
General-purpose robots require decision-making models that generalize across diverse tasks and environments. Recent works build robot foundation models by extending multimodal large language models (MLLMs) with action outputs—creating vision-language-action (VLA) systems. These efforts are motivated by the intuition that MLLMs' large-scale language and image pretraining can be effectively transferred to the action output modality.
%
In this work, we explore an alternative paradigm of using large-scale video pretraining as a primary modality for building robot foundation models.
Unlike static images and language, videos capture spatio-temporal sequences of states and actions in the physical world that are naturally aligned with robotic behavior.
%
We curate an internet-scale video dataset of human activities and task demonstrations, and train, for the first time at a foundation-model scale, an open video model for generative robotics planning. The model produces zero-shot video plans for novel scenes and tasks, which we post-process to extract executable robot actions. We evaluate task-level generalization through third-party selected tasks in the wild and real-robot experiments, demonstrating successful physical execution. Together, these results show robust instruction following, strong generalization, and real-world feasibility. We release both the model and dataset to support open, reproducible video-based robot learning. Our website is available at~\href{https://www.boyuan.space/large-video-planner/}{https://www.boyuan.space/large-video-planner/}.


\end{abstract}
\section{Introduction}


\looseness=-1
A key component of many robotic systems is the planning algorithm~\citep{garrett2021integrated}, which takes a high-level task or instruction alongside the robot's sensory observations to generates a sequence of states and actions that will achieve the goal. General-purpose robots — systems designed to operate reliably across diverse tasks and novel environments — would greatly benefit  from planning algorithms that are themselves extremely general. Such planning algorithms should be able to comprehend unseen tasks, adapt fluidly to novel scenes, and output physically coherent behaviors. Developing these strong generalization capabilities remains a central, unresolved challenge of embodied intelligence today.

The recent success of foundation models in language and vision has reshaped how generalization is achieved in AI. Large language models (LLMs)~\citep{achiam2023gpt,bai2023qwen, touvron2023llama, comanici2025gemini, liu2024deepseek, team2025kimi} trained on internet-scale text corpora exhibit broad competence across unseen tasks, suggesting that scale and data diversity can induce powerful transfer. Extending this idea, multimodal large language models (MLLMs)~\citep{liu2023visual, team2024chameleon, bai2025qwen2, han2025learning} align vision and language, grounding textual reasoning in perception. These advances have inspired robot foundation models—large, unified architectures that generalize across scenes and tasks by integrating perception, language, and control. A key instantiation is the Vision–Language–Action (VLA) model~\citep{brohan2022rt, kim2024openvlaopensourcevisionlanguageactionmodel, black2024pi_0}, which extends MLLMs with an action output modality. 

\looseness=-1
However, in comparison to web-scale text and image data that underpin MLLMs~\cite{schuhmann2022laion, penedo2024fineweb, chen2024panda}, robot action data is much scarcer~\citep{open_x_embodiment_rt_x_2023, bu2025agibot}.  As a result, it is difficult to build VLA models with the same level of competency as that of MLLMs, with existing VLAs relying on an asymmetric form of transfer, where the pretrained knowledge in an MLLM is finetuned on a narrow amount of robot data. Such a construction leads to poor generalization when given new robot tasks in unseen situations~\citep{pi0-experiment-wild}.


In this work, we propose an alternative paradigm for robot foundation models—using video as the primary modality. Unlike static image–text pairs, videos naturally encode state–action plans, visually depicting how the world evolves as agents interact with it. A video generative model conditioned on a textual instruction and an initial observation frame can predict plausible future frames—effectively generating visual action plans for diverse tasks. This formulation aligns closely with robotics: it captures spatial and temporal continuity, offering a far richer representation of continuous actions than text tokens. Moreover, video data is abundant online, spanning human activities, instructional tutorials, and task demonstrations. Each video implicitly contains action information, following the same pretraining principle that powers MLLMs—leveraging large, naturally occurring data to learn mappings grounded in real-world behavior. Compared to the asymmetric transfer of VLAs, this video-based paradigm offers a directly grounded source of transfer: the data itself captures the temporal dynamics of action, providing a stronger bridge to downstream embodied tasks.

\begin{figure*}[t]
    \centering
    \includegraphics[width=\linewidth]{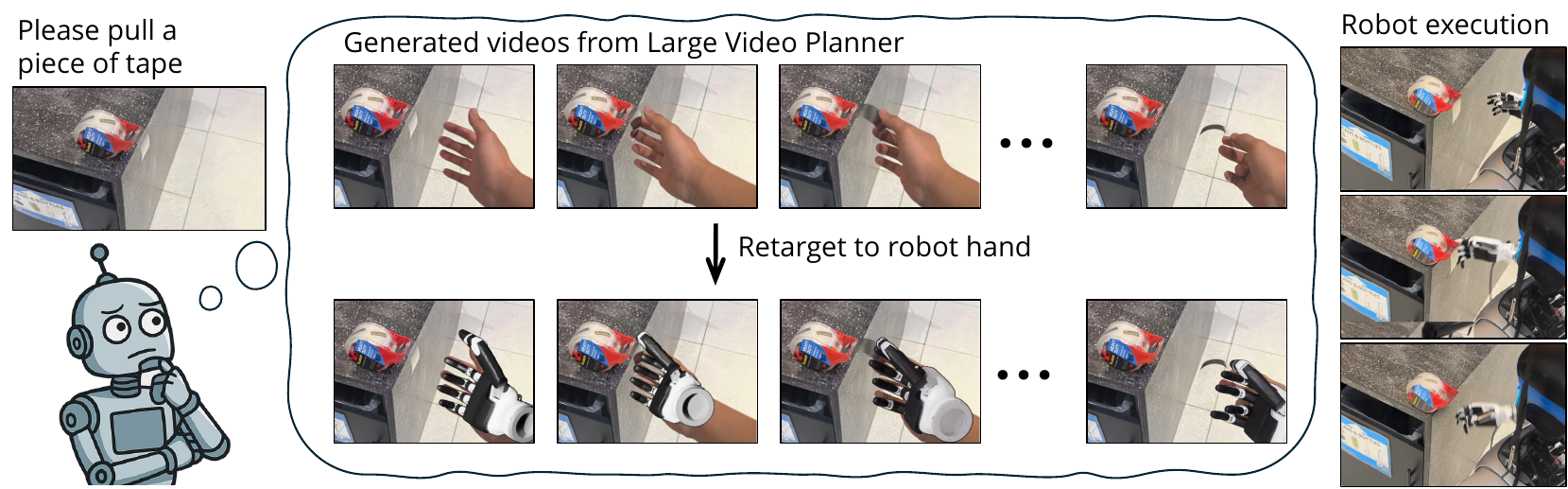}
    \vspace{-15pt}
    \caption{\textbf{Autonomous Robot Execution with Large Video Planner.} Our approach uses video generation as a visual motion planner in pixel space. From a single image and a task instruction, the model generates a video depicting how the task should be completed. The predicted human motion is then retargeted to a robot hand for real-world execution, enabling zero-shot visual planning in diverse scenes. }
    \vspace{-5pt}
    \label{fig:teaser}
\end{figure*}

To instantiate this paradigm, we develop a video foundation model purpose-built for embodied decision-making and action extraction. Unlike existing video generation models—typically optimized for content creation and prone to mode collapse—we prioritize the emphasizes on
physical consistency to real image observations and adherence to task instructions. It takes textual instructions and initial observation frame(s) as input, generating predictive video plans from which executable actions are extracted~\citep{du2023learning}. 

We realize this vision through advances in both data and model design. 
First, we introduce and release a large-scale, open dataset curated for embodied decision making.
Sourced from a diverse mix of internet and robot videos, this dataset is carefully processed to capture complex human and robot behaviors, maintain high temporal coherence, and ensure tight alignment with language instructions.
Second, we propose a novel model that leverages History Guidance~\citep{song2025historyguidedvideodiffusion} and Diffusion Forcing~\citep{chen2024diffusion} to  specifically enhance temporal coherence and causal consistency in the generated frames.
These data and model contributions jointly enable large-scale, temporally grounded pretraining, leading to robust generalization across diverse tasks and scenes.

We evaluate our model’s generalization through two complementary experiments. First, we perform an extensive independent evaluation to assess task-level generalization of the video model itself: independent testers freely selected scenes and tasks—encouraged to be creative and challenging—producing evaluations spanning diverse conditions, from outdoor scenes like crosswalks to dexterous tasks such as tearing tape. Second, we conduct real-robot experiments demonstrating end-to-end execution. Actions extracted from generated video plans are deployed on physical robots, from parallel grippers to dexterous hands, successfully executing tasks in uncontrolled environments. Compared to baseline models, our approach exhibits a stronger grasp of contact dynamics, goal reasoning, and end-to-end execution, demonstrating robust generalization across both simulated and real settings.

In summary, our work makes three primary contributions. (1) Large Video Planner (LVP), a large-scale video foundation model designed for robotic manipulation, and an associated framework for deploying it as a zero-shot policy on real robots. (2) A curated, open internet-scale video dataset of human activities and robot task demonstrations, carefully processed for embodied decision making and instruction following. (3) A rigorous evaluation of task-level generalization, using an independent testing protocol and real-robot experiments to systematically assess generalization across unseen environments, tasks, and embodiments.

\section{Related Work}

\textbf{Video Diffusion.}
Diffusion models~\citep{sohl2015deep, ho2020denoising, lipman2022flow} currently represent the state of the art for synthesizing high-quality videos~\citep{imagenvideo2022, svdblattmann2023, sora2024, wan2025}. They generate videos by iteratively denoising an entire sequence of frames jointly, often operating in a lower-dimensional latent space for efficiency.  Recent works adapt this process to better capture the temporal structure, yielding improved results over vanilla full-sequence diffusion~\cite{rolling_diffusion, chen2024diffusion, yin2025slow, huang2025self}. Rolling Diffusion~\citep{rolling_diffusion} introduces a sliding-window diffusion process that progressively corrupts data from past to future, enabling autoregressive rollouts for long video generation. Diffusion Forcing~\citep{chen2024diffusion} generalizes this idea by training with independently randomized noise levels across tokens, allowing the model to generate sequences in arbitrary order and with next-token or next-few-token denoising.

To improve visual conditional generation, classifier-free guidance~\cite{ho2022classifier} has proven to be an essential technique to improve fidelity and controllability. It is mostly applied with text based conditional signals. History Guidance~\citep{song2025historyguidedvideodiffusion} extends this paradigm by conditioning generation on one or several preceding frames, amplifying temporal grounding. This approach has been applied in tasks such as image-to-video generation and video extension~\citep{zhangframe, song2025historyguidedvideodiffusion}, where temporal grounding is critical.

\textbf{Robot Foundation Models.}  
A large body of recent work has focused on constructing robot foundation models from large-scale robotics datasets~\citep{ open_x_embodiment_rt_x_2023, khazatsky2024droid, ebert2021bridgev1, walke2023bridgev2,lirose, geng2025roboverseunifiedplatformdataset}.  
Such models include \emph{Vision–Language–Action (VLA) policies}~\citep{brohan2023rt2visionlanguageactionmodelstransfer, octomodelteam2024octoopensourcegeneralistrobot, kim2024openvlaopensourcevisionlanguageactionmodel, gr00tn1_2025, black2024pi0visionlanguageactionflowmodel,team2025gemini, team2025gemini}, which directly map multimodal inputs to robot actions, as well as \emph{large diffusion and transformer-based networks} that learn to generate trajectories or action sequences~\citep{zhu2025scaling, barreiros2025careful}.  
Another line of work\cite{ahn2022icanisay, geng2023sage, open6dor, li2023manipllm, huang2023voxposercomposable3dvalue} explores \emph{embodied LLMs and planner–policy hybrids}, which leverage large language models for high-level reasoning and planning while delegating execution to low-level controllers, e.g., PaLM-E~\citep{driess2023palmeembodiedmultimodallanguage} and RoboFlamingo~\citep{li2023vision}.  

In contrast to these approaches, we formulate a robot foundation model as a \emph{video generation model} that produces realistic interaction videos depicting what the robot should do, from which executable actions can then be extracted and retargeted.

\textbf{Video Generation for Robotics.}
%
%
Learning from video demonstration has been an interesting research direction for robot learning~\citep{liu2018imitation, learning_by_watching, shao2021concept2robot,mccarthy2025towards}, partially due to its richness of data. With recent advances in video generation, researchers have begun exploring its use in robotics. One line of work employs video generation for visual policies, where models synthesize videos of successful task completion to guide control~\citep{du2023learning, liang2024dreamitate, bharadhwaj2024gen2act, luo2025solving, agarwal2025cosmos}. Another line treats video generation as a dynamics model, predicting future frames conditioned on action inputs~\citep{ajay2023compositional, yang2023learning, du2023video, zhu2024irasim, qi2025strengthening}. Video world models have also been used as evaluators, with the potential of enabling robots to validate and refine their strategies more efficiently through simulated rollouts~\citep{monas_jang_1x_world_model_2024}. The distinction between policy learning and dynamics modeling is increasingly blurred, as recent approaches train unified models that jointly generate both videos and robot actions~\citep{li2025unified}.

Given a generated video, there are several approaches to extract continuous actions. One approach is to learn an inverse dynamics or goal-conditioned policy on top of the generated videos~\citep{du2023learning, wang2025language, hong2024slowfast, luo2024grounding, zhen2025tesseract}. Alternatively, we can directly infer 2D or 3D scene flow to obtain actions~\citep{ko2023learning,chen2025vidbot}. In contrast, in this paper we illustrate how we can leverage 3D scene reconstruction MegaSam~\citep{li2024megasamaccuratefastrobust} followed by hand reconstruction using HaMeR~\citep{hamer} and Dex-Retargeting~\citep{qin2023anyteleop} to obtain actions.

\section{Method}
A robot foundation model maps observations and goals to a sequence of actions. We realize this through a two-stage design: a large video planner followed by action extraction. Consider a robot facing a door it has never encountered before. Its camera perceives the door handle as its owner instructs, ``Open this door.'' The robot first employs a video foundation model to imagine how a rational human would perform the task—generating a video where a hand reaches for the handle, twists it, and pushes the door open. It then applies action extraction algorithms to translate this visual plan into executable control signals, whether for a dexterous five-fingered hand or a parallel gripper.

In Section~\ref{subsubsec:video_foundation_model}, we present our  video foundation model for generative planning in video space. Section~\ref{subsubsec:data} introduces an internet-scale video dataset of human activities and robot demonstrations that we curated for this study. Finally, Section~\ref{subsec:video2robot} describes our retargeting mechanism that grounds the generated video plans into robot-specific actions across diverse morphologies.

\subsection{LVP: A Video Foundation Model for Generative Planning}

\label{subsubsec:video_foundation_model}

\paragraph{Latent Diffusion} We begin building our video foundation model following the latent diffusion framework~\citep{sora2024, wan2025}. We use a temporally causal 3D variational autoencoder (VAE) to compress a video clip in pixel space into a compact, lower-dimensional latent representation $x$. The VAE encodes each  $8\times 8 \times 4$ spatiotemporal patch into a 16-channel embedding, converting an input of shape $[1+T, 3, H, W]$ into a latent of shape $[1+\lceil T/4 \rceil, 16, \lceil H/4 \rceil, \lceil W/4 \rceil$, where $T+1$ is the number of frames and $H, W$ are spatial dimensions. The first frame of a video is repeated $4$ times before such compression to allow co-training with single-frame image data, which corresponds to the $1$ in $T+1$.

\begin{figure*}[t]
    \centering
    \includegraphics[width=\linewidth]{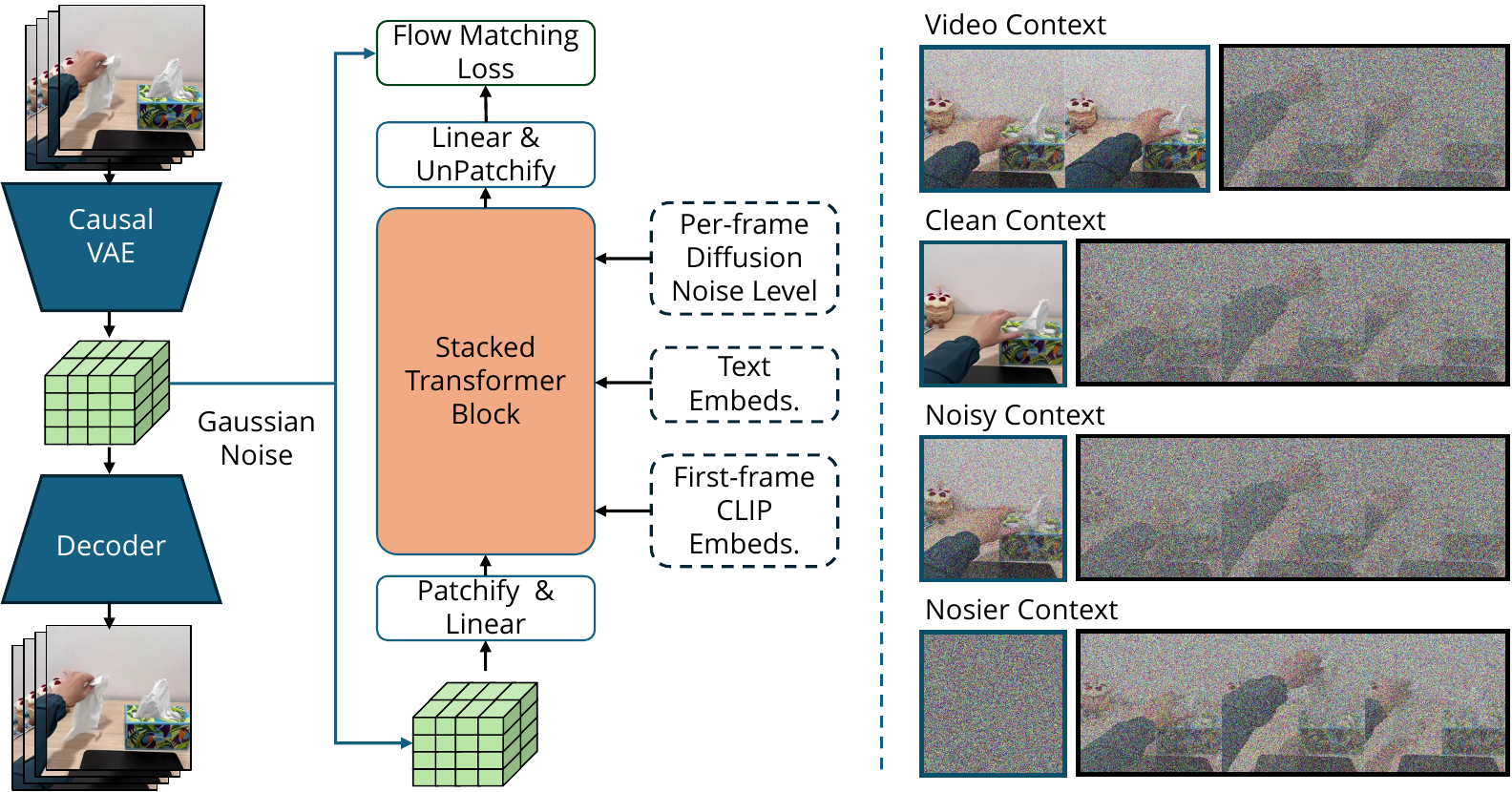}
    \put(-346, -8){\capfont \textbf{(a)} Latent video diffusion}
    \put(-125, -8){\capfont \textbf{(b)} Diffusion forcing  training}
    \caption{\textbf{LVP Overview:} \textbf{(a)} Overview of the latent video diffusion framework. We first use a temporally causal VAE to encode video clips into compressed 3D latent representations. Then we train a diffusion transformer in this latent space with flow matching objectives.   \textbf{(b)} We jointly train image-to-video (I2V) and video-to-video (V2V) with a modified diffusion forcing training strategy. During training, a random context length between $0$ and $6$ frames is selected, dividing the video into history and future segments. Two independent noise levels are applied to these segments, and the history segment is set to zero noise with a 50\% probability. We visualize four representative cases of this noisy training strategy: the top row shows that longer contexts enable V2V training; the second row shows clean first-frame contexts, which exactly aligns with standard I2V training; and the botton two rows show noisy context frames, which improve robustness to out-of-distribution conditioning.}
    \label{fig:video_diffusion_arch}
\end{figure*}

We then freeze the 3D VAE and train a special video diffusion model~\citep{sohl2015deep, ho2020denoising, lipman2022flow, peebles2023scalable} in this compressed latent space using a modified Diffusion Forcing Transformer~\citep{chen2024diffusion,song2025historyguidedvideodiffusion}, a DiT~\citep{peebles2023scalable} variant we introduce below, and illustrated in Figure~\ref{fig:video_diffusion_arch}. Following the diffusion training recipe, we add Gaussian noise to a clean video latent and train our diffusion model to remove such noise. At sampling time, starting from a latent pre-filled with noise, the model iteratively denoises the latent until obtaining a clean sample. The VAE decoder then decodes this latent into a video sample.

Specifically, we train this video diffusion model with the flow matching objective~\citep{lipman2022flow}. Under a shifted schedule~\citep{esser2024scaling} that emphasizes higher noise levels, we add noise to an encoded video latent $z_0$ by $z_k = (1 - k) z_0 + k \epsilon$, where $k$ denotes the chosen noise level, $\epsilon\sim\mathcal{N}(0,1)$ and $z_k$ is the noisy latent.
The model $f_\theta$ is trained to predict the flow $\epsilon - z_0$ conditioned on the noisy latent $z_t$, conditioning $c$ (comprising the input image and text instruction), and noise level $t$, minimizing the matching loss~\citep{lipman2022flow} $\mathcal{L} = ||f_\theta(z_k, c, k) - k(\epsilon - z_0) ||_2$.


\paragraph{Diffusion Forcing Transformer}
\label{sec:dfot}
A challenge in the video diffusion model is temporal coherence. In our formulation, the generated video must be coherent with not only the language instruction (text-to-video or t2v) but also the first frame (image-to-video or i2v) specified by the robot observation. In addition, one may want to condition the video generation on multiple previous frames (video-to-video or v2v) to generate multi-stage video plans. Traditionally, one achieves such conditioning by finetuning a t2v model to cross-attend to separate patches of the context frame(s)~\citep{wan2025}. We, however, propose to better satisfy this need with the recently proposed Diffusion Forcing framework.

Instead of adding a uniform level of noise to all tokens like in legacy video diffusion models, Diffusion Forcing~\citep{chen2024diffusion} found that training video diffusion models with different noise levels at different frames has the additional benefit of flexibility and rollout stability. Since all the noise levels are random during training, at test time one can flexibly control the conditioning by selecting the desired noise level.


To learn i2v and v2v with a unified objective, we adopt diffusion forcing and apply different noise levels to context frames versus generated frames. As shown in Figure~\ref{fig:video_diffusion_arch}(b), given a diffusion transformer on a fixed number of latent frames, we first randomly sample a history length from $\{0, 1,2, \ldots, 6\}$ latent frames, splitting the video into a history segment and a future segment. We then apply independent noise levels to each segment and feed the resulting noisy video to our model, leaving all other settings unchanged.  For example, if one adds zero noise to the first frame or first few frames at training time, the model will find it as a perfectly visible context frame and learn to condition on it; if the history frames have an intermediate noise level, the model treats it as partial information and learns to be robust to out-of-distribution context frames. In this way, we can flexibly condition on a clean first frame or multiple history frames at sampling time, by setting the their noise levels to 0.

Not only does this method eliminate an extra cross-attention to variable-length context tokens, but it's also compatible with existing DiT model weights without architectural changes. Following~\citet{song2025historyguidedvideodiffusion}, we simply feed different noise level embeddings to different tokens in the DiT architecture, instead of uniform ones. This allows us to train a Diffusion Forcing model on top of the weights of a pre-trained video foundation model, WAN 2.1 14B~\citep{wan2025}. Following the practice of WAN 2.1 14B, we cross-attend to the CLIP features of the first frame as well as the text embeddings extracted by the UMT5~\citep{chung2023unimax} encoder. Because Diffusion Forcing achieves context frame conditioning in a cleaner way, we remove WAN's mask and guidance channels used for image conditioning.

\paragraph{Enhanced Temporal Coherence with History Guidance}
In addition to flexible conditioning and compatibility with legacy weights, our design can significantly enhance context coherence by enabling special sampling techniques from Diffusion Forcing ~\citep{chen2024diffusion,song2025historyguidedvideodiffusion}.

Classifier-Free Guidance (CFG)~\citep{ho2022classifier} is known to improve visual quality and conditioning adherence in visual generative models. WAN 2.1 utilizes a text-CFG that combines the output of a text-conditional diffusion model and that of an unconditional one. However, this still yields unsatisfactory motion fidelity and weak image conditioning as shown in Figure~\ref{fig:video_results}.

LVP adopts \textit{history guidance}~\citep{song2025historyguidedvideodiffusion}, a CFG variant that performs guidance on any amount of context frames. Let $x_k$ denote the future segment to be diffused at noise level $k$ and $c_{\text{text}}$ the task instruction. As our model is trained with Diffusion Forcing, we can flexibly condition on a provided history segment $x_{\text{hist}}$ at sampling time by setting its noise level to zero, be it a single frame or a context video:
\begin{equation}
    \nabla \log{p(x_k | x_{\text{hist}}, c_{\text{text}}, k)}.
\end{equation}
Similarly, we can set the noise level of context frames to the maximum to fully mask out the context frames and obtain the unconditional score:
\begin{equation}
    \nabla \log{p(x_k | c_{\text{text}}, k)}.
\end{equation}
To perform history guidance, we sample with the combined score
\begin{equation}
\begin{aligned}
    s_{\text{hist}}  &= (1+w_{\text{hist}})\nabla \log{p(x_k | x_{\text{hist}}, c_{\text{text}})}  - w_{\text{hist}}\nabla \log{p(x_k | c_{\text{text}})}. \\
\end{aligned}
\end{equation}
Just as text-based CFG enhances adherence to text instruction, history guidance enhances adherence to context images. During sampling, we combine both history guidance and text-based CFG to generate videos that adhere to both text and context frames by sampling with the score
\begin{equation}
\begin{aligned}
s_{\text{final}} = & (1+w_{\text{hist}})\nabla \log{p(x_k | x_{\text{hist}}, c_{\text{text}})}  - w_{\text{hist}}\nabla \log{p(x_k | c_{\text{text}})}+\\
& (1+w_{\text{text}})\nabla \log{p(x_k | x_{\text{hist}}, c_{\text{text}})}  - w_{\text{text}}\nabla \log{p(x_k | x_{\text{hist}})}
\end{aligned}
\end{equation}

This combined guidance technique can significantly enhance the plan quality compared to traditional text-based guidance, yielding physically viable plans with strong instruction following.

\paragraph{Autoregressive Extension for Multi-Stage Planning}  Due to the flexible history conditioning, our model can extend a previously generated or captured video. The model supports up to 24 frames (6 latent frames in VAE space) as context. We can repeat video extension iteratively to generate multi-stage video plans. See Figure~\ref{fig:video_multistage} and the videos on our website for multi-stage results.

\paragraph{Training Details}
%
%
%
%
We train the model in two stages to progressively improve its visual planning capability and visual quality:
\begin{itemize}[leftmargin=2em, itemsep=0.05em]
\vspace{-5pt}
    \item \textbf{Continue pretraining}. Starting from Wan I2V 14B weights, we discard the weights that handle the extra masking and image guidance channels. We train on the full dataset for 60k steps with a batch size of 128, for a total of 200B tokens. At this stage, the model captures rich dynamics and strong instruction-following behavior, but the generated videos often exhibit excessive camera motion, which hinders smooth deployments on robots.
    \item \textbf{Low camera motion finetuning}. To reduce unwanted camera motion, we curate a smaller subset from Ego4D, Epic-Kitchens, and Panda datasets by selecting clips with a much lower average optical flow magnitude, and finetune for an additional 10k steps. This stage effectively suppresses camera drift and improves overall temporal smoothness and visual stability.
\end{itemize}

The total training takes around 14 days with 128 H100 SXM5 GPUs.

\begin{figure*}[t!]
    \centering
    \includegraphics[width=\linewidth]{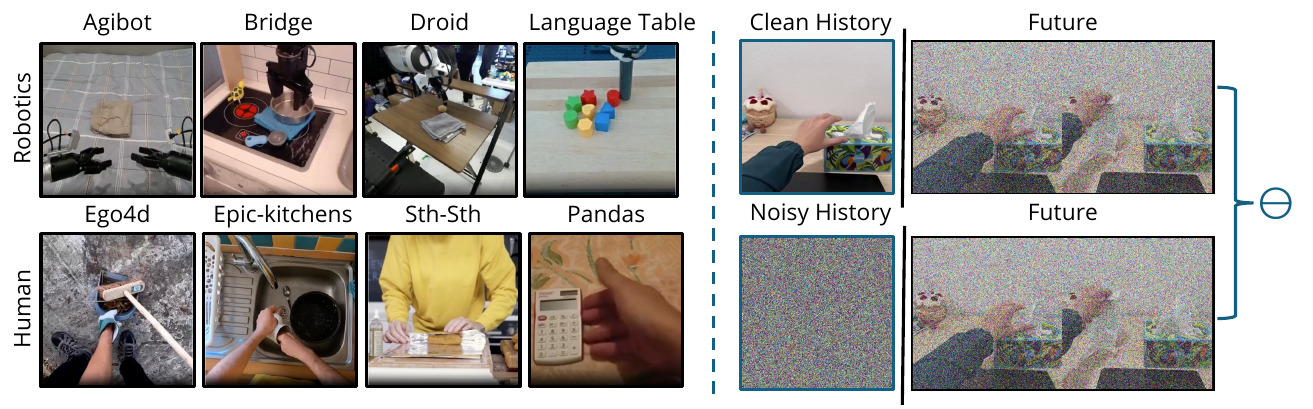}
    \put(-346, -5){\capfont \textbf{(a)} Dataset source visualization}
    \put(-155, -5){\capfont \textbf{(b)} History-guided sampling}
    \caption{ \textbf{(a)} Visualization of our eight dataset sources. First row: four robotics datasets. Second row: four human-centric datasets. \textbf{(b)} Illustration of our video diffusion sampling strategy, where scores estimated with and without history are linearly combined. Text conditioning and the diffusion transformer are omitted for clarity.}
    \label{fig:video_diffusion}
\end{figure*}

\subsection{\dataname: A video dataset of human and robot actions}
\label{subsubsec:data}
Training a video foundation model for \textit{embodied planning} demands abundant data emphasizing diverse object interactions with action-centric text annotations. This contrasts with standard video datasets used for content-creation-oriented video generation \citep{kong2024hunyuanvideo, wan2025}, which often prioritize aesthetic quality, cinematic shots, or dense captions of visual appearance and elements rather than motions. To this end, we curate \textbf{\dataname{}}, a diverse and high-quality dataset of 1.4M short clips showing humans or robots interacting with objects, each paired with multiple action-centric captions.



\paragraph{Video Sources} 
Given the vast availability of video data, we source raw videos from existing datasets before providing our own high-quality annotations. To ensure broad diversity across scenes, tasks, and embodiments, we combine robot teleoperation and human activity videos. 

We start with web crawls widely used by video foundation models. We choose Pandas 70M~\citep{chen2024panda} as a source of videos for heavy filtering. These internet-scale datasets contain diverse videos filtered for visual quality and captioned with visual content. They provide crucial scale and diversity that span countless tasks, scenes, and objects. However, only a small proportion of these videos capture detailed hand interactions with objects at sufficient resolution, not to mention near-zero robot coverage. 

A second source of video comes from egocentric human activity datasets. These medium-scale datasets contain many annotated human–object interactions with moderate diversity but often suffer from large background motion due to camera movement. In addition, we found that atomic action annotations in these datasets still have lengths varying from seconds to minutes. We opt to draw videos from Ego4D~\citep{grauman2022ego4d}, Epic Kitchens~\citep{Damen2022RESCALING}, and Something-something~\citep{goyal2017something} dataset before heavy filtering, frame alignment and recaptioning.

A third source of videos come from robotics datasets featuring teleoperated robots performing tasks.  While they provide knowledge about robot morphology, spanning parallel-jaw grippers to dexterous multi-fingered hands, they often have poor visual quality, poorly aligned frame rates, and limited diversity. Further, we found that short captions describing the task are often lacking, with many videos annotated with captions as vague as ``pick'', or not featuring a caption at all. We opt to select Bridge~\citep{walke2023bridgev2}, Droid~\citep{khazatsky2024droid}, Language Table~\citep{lynch2023interactive}, and AgiBot-World~\citep{bu2025agibot} for heavy captioning and frame alignment.

We provide a summary of video sources and their key properties in Table~\ref{tab:dataset_collections}, examples from each dataset in Figure~\ref{fig:video_diffusion}(a), and additional details in App.~\ref{app:data_details}. Together, we hope our model will achieve synergy by learning better instruction following from the diversity of web crawls, better object-hand interactions from egocentric human activities, as well as robotic morphologies from robot data.

\paragraph{Temporal Alignment} We train our model to generate 3-second action videos at 16 frames per second, as this provides a good trade-off between computational cost and action granularity. However, we observe that different datasets often feature varying-length clips for atomic actions, spanning 1 second to 1 minute.  A closer examination reveals that robotics datasets contain motions much slower than those of humans performing the same tasks and are often recorded at drastically different frame rates, sometimes as low as 5 fps.

Rather than naively aligning frame rates or trimming a video clip to a target length, we deem it important to align all clips to human speed to avoid temporal inconsistency - if a human normally finishes the task in 3 seconds, we resample the robot video (via upsampling or speeding up) so that it performs the same task in 3 seconds, regardless of its original frame rate or teleoperation speed. We achieve this by visually inspecting all datasets to determine the appropriate subsampling ratio following this principle. We also break down long-horizon tasks into atomic actions if any annotation contains multi-stage tasks. Some egocentric human activity datasets already provide action clip annotations, but we further refine them by trimming each clip precisely at the action’s start and end points. As we found later in experiments, such alignment is critical to enhancing the transfer between different morphologies.

\paragraph{Quality Filtering} 
After temporal alignment, we first discard clips that are low-resolution, too short, too long, or poorly lit. We then apply some additional filters to focus the model on embodied motion planning: 

\textit{Filtering rapid camera motions}. Many egocentric videos exhibit rapid camera rotations, leading to large background shifts and high training loss. These distract the model from learning meaningful foreground object motions. To mitigate this, we filter videos using optical flow statistics. 
 
 
 \textit{Ensuring visible embodiment}. To avoid ambiguity, we require the embodiment (hand or robot gripper) to be clearly visible in the first frame. We use object detectors to automatically filter out clips where the embodiment is absent.

\textit{Expert motion}.
Many robot datasets contain suboptimal trajectories where the robot does not successfully accomplish the task. Traditionally, robot foundation models do not filter such data even when a “success” annotation is provided. However, we consider it important to remove these failure trajectories.

\textit{Filtering Pandas-70M subset.}
We perform three stages of progressive filtering to extract a subset focused on human interactions from the large Panda-70M dataset. First, we perform keyword-based filtering on captions using a whitelist (e.g., ``grasping'', ``pull'') and a blacklist (e.g., "cartoon," "video game"). We then use human detectors to retain only clips containing one to four humans visible in the first, middle, and last frames. Finally, we perform another round of filtering with Gemini. For each video clip, we prompt Gemini with four questions to verify whether the clip contains rich human hand motions. We list more details of this in the Appendix~\ref{app:subsec_pandas}.

\paragraph{Action-Centric Re-Caption} 
To enhance the instruction following of our model, we generate multiple high-quality captions for each video. Traditionally, video foundation models favor extremely detailed captions describing all the visual elements. We observe that some robot datasets or ego-centric datasets only feature extremely short task descriptions as simple as one word, such as ``pick''. For these videos (e.g., DROID, Ego4D), we prompt Gemini Flash with the instruction and initial frame to create more detailed and varied captions. For videos that lack task annotations, we prompt Gemini with the entire video clip and ask it to describe the primary action and involved objects. 

In total, including repeated captions, we obtain 4.1 million captioned clips. We find that providing Gemini with a short description (e.g., ``pick up a cup”) substantially improves caption accuracy, as the video perception of Gemini tends to describe static scene contents well but struggles with fine-grained action dynamics. We make sure each video clip is paired with two to five distinct captions, some short, some extremely descriptive, to enhance linguistic diversity and improve training robustness.

\begin{table}[h!]
\centering
\resizebox{\columnwidth}{!}{%
\begin{tabular}{llccccc}
\toprule
\textbf{Name} & \textbf{\# Filtered clips} & \textbf{Robot?} & \textbf{Ego-centric?} & \textbf{Morphology} & \textbf{In-the-wild} & \textbf{Bimanual} \\
\midrule
Bridge & 25k & Yes & Third-person & Gripper & No & No \\
DROID & 192k & Yes & Third-person & Gripper & No & Yes \\
Language-Tables & 71k & Yes & Third-person & Gripper & No & No \\
AgiBot-World & 863k & Yes & Third-person & Gripper & No & Yes \\
Ego4D & 39k & No & Egocentric & Human Hand & Yes & Yes \\
Epic-Kitchens & 7k & No & Egocentric & Human Hand & No & Yes \\
Something-Something & 93k & No & Third-person & Human Hand & Yes & Yes  \\
Panda-70M (filtered) & 196k & No & Third-person & Human Hand & Yes & Yes \\
\midrule
\bottomrule
\end{tabular}%
}
\vspace{-5pt}
\caption{\textbf{\dataname} sources and properties after curation. The dataset targets action-centric clips with broad diversity across embodiment (robot/human), viewpoint (ego/third-person), scene type (in-the-wild/lab), morphology, and bimanuality, totaling 1.4 million clips.}
\label{tab:dataset_collections}
\end{table}

\subsection{Robot Actions from Video Plans}  
\begin{figure*}[t]
    \centering
    \includegraphics[width=\linewidth]{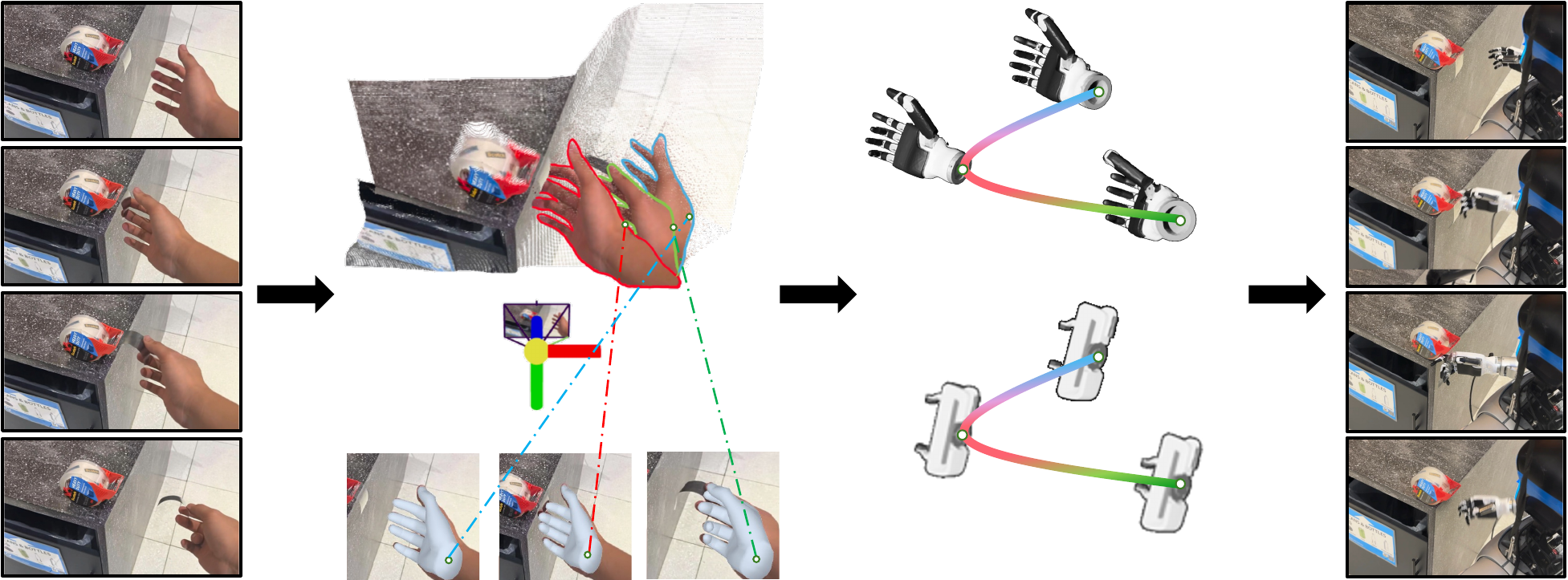}
    \put(-400, -9){\capfont Generated videos}
    \put(-290, -9){\capfont Hand pose estimation}
    \put(-290, 36){\capfont 4D reconstruction}
    \put(-190, -9){\capfont Wrist motion \& finger retargeting}
    \put(-58, -9){\capfont Robot execution}
    \caption{\textbf{Pipeline from Video to Action}.  Given a generated video depicting a human hand performing a task, we first reconstruct and track the hand in 3D (second column). The reconstructed hand motion is then retargeted to dexterous hands or grippers (third column). Finally, the retargeted trajectory is transformed into the robot’s control frame and executed in the real world (rightmost column). }
    \label{fig:placeholder}
\end{figure*}
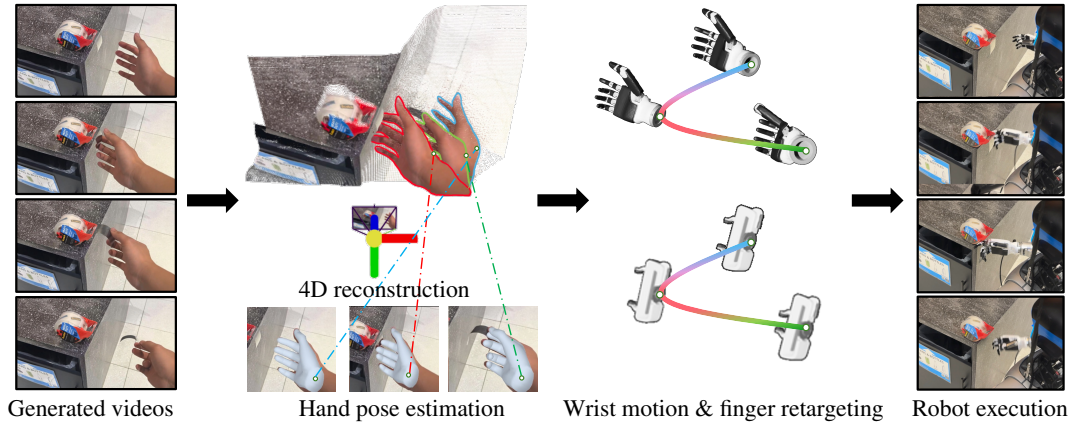

\label{subsec:video2robot}




Given a camera observation and a task description, our large video planner can generate a video plan of a human hand or robot gripper executing the task. This section describes how we extract executable actions from a video plan and deploy them on robots. 

Our action extraction pipeline supports retargeting generated human hand video to a dexterous hand or even a simple robot gripper (see Appendix~\ref{appendix: gripper}). In this section, however, we primarily focus on one type of transfer: human hand video to dexterous robot hand execution, as the majority of our robot experiments are done with a humanoid robot with a dexterous hand.

\paragraph{Human Hand Motion Estimation} We reconstruct an accurate and temporally aligned hand pose as first step for motion retargeting. 
To do so, we first predict hand pose  in each video frame independently using image-based hand reconstruction model:\emph{HaMeR} \citep{hamer}, then align and refine the predicted human hand with a dynamic scene reconstruction model, \emph{MegaSAM}~\citep{li2024megasamaccuratefastrobust}. 


\textit{Per-frame Hand Pose Estimation.}
For each input frame $I_t$, \emph{HaMeR} predicts MANO~\citep{romero2022embodied} hand vertices $\mathbf{V}_t$ and a global wrist orientation $\mathbf{R}_t \in \mathrm{SO}(3)$ in the camera coordinate frame. While HaMeR provides accurate hand shape and articulation, its per-frame translation estimates tend to drift over time due to the lack of temporal consistency enforcement. 



\textit{4D Consistent Alignment.}
We then align the translations of the per-frame reconstructed human hand. Specifically, we leverage a 4D reconstruction model, \emph{MegaSAM}~\citep{li2024megasamaccuratefastrobust}, which outputs per-frame depth maps $D_t(u,v)$, camera intrinsics $\mathbf{K}$, and extrinsics $\{\mathbf{E}_t\}_{t=0}^{T-1}$.  After getting per-frame depth and camera pose, we backproject pixels of the hand into 3D, where pixels of the hand $(u_t, v_t)$ are obtained by projecting the MANO wrist joint regressed from $\mathbf{V}_t$. 

We retain HaMeR’s orientation $\mathbf{R_t}$  while using the backprojected wrist pointclouds to estimate $\mathbf{T_t}$. This enforces temporal smoothness, resolves monocular scale ambiguity, and significantly improves wrist localization robustness.

\textit{Temporal Completion and Smoothing.}
Frames with invalid depth/pixels are marked missing and linearly interpolated in position. Quaternions use SLERP with sign flips to maintain continuity. We then apply a causal Savitzky-Golay filter (window $w$, order $d$) to positions and quaternion components, followed by re-normalization, noted as $\hat{\mathbf{T}}_{\mathcal{R}\leftarrow\mathcal{W},t}$.

\paragraph{Robot Finger Motion Retargeting}
Given the human hand pose estimated by the previous module, we design retargeting modules that support both multi-finger dexterous hands and parallel-jaw grippers. We introduce multi-finger dexterous hands below and parallel-jaw grippers in Appendix~\ref{appendix: gripper}.

To retarget robot finger joints from human hands, we use \emph{Dex-Retargeting}~\citep{qin2023anyteleop}, which first extracts human hand keypoints using an RGB-based detector and then maps them to robot joint configurations by solving a DexPilot-style optimization objective. This produces robot finger joint angles $\mathbf{q}^R_t \in \mathbb{R}^{n_{\mathrm{dof}}}$, enabling fine-grained imitation of articulated human manipulation.


We export per-frame wrist SE(3) $\big(\hat{\mathbf{p}}^{\mathcal{R}}_{\mathcal{W},t},\hat{\mathbf{q}}^{\mathcal{R}}_{\mathcal{W},t},\hat{\mathbf{T}}_{\mathcal{R}\leftarrow\mathcal{W},t}\big)$ and robot joints $\{\mathbf{q}^R_t\}$, together with metadata (joint names, DOF). Qualitative checks are performed by rendering the robot hand motions in simulation, see videos in the project website.

\paragraph{Real-Robot Execution} Given the human wrist trajectories $\{\mathbf{P}_{t}\}_{t=0}^{T-1}$ and the robot finger joint trajectories $\{\mathbf{q}_{t}\}_{t=0}^{T-1}$ estimated by the preceding modules (both expressed in the camera coordinates of the first video frame), our goal is to execute the motion on a physical robot. We first rotate the wrist poses into the robot control frame, then use the resulting wrist translations and orientations to solve the inverse kinematics (IK) 
for the arm (using cuRobo\cite{curobo_report23} ), while the finger trajectories directly drive the robot hand joints.

\textit{Camera-to-Robot Alignment.}
Let $\mathcal{C}_0$ denote the coordinate frame of the first camera, $\mathcal{M}$ the MANO hand frame, and $\mathcal{R}$ the robot control frame. We align the recovered wrist poses from $\mathcal{C}_0$ to $\mathcal{R}$ via an extrinsic calibration. In practice, this reduces to applying a fixed rotation $\mathbf{M}\!\in\!\mathrm{SO}(3)$ that unifies the axes of $\mathcal{C}_0$ and $\mathcal{R}$:
\begin{equation}
\label{eq:frame-map}
\mathbf{p}^{\mathcal{R}}_{\mathcal{W},t} = \mathbf{M}\,\mathbf{p}^{\mathcal{C}_0}_{\mathcal{W},t} + t,
\qquad
\mathbf{R}^{\mathcal{R}}_{\mathcal{W},t} = \mathbf{M}\,\mathbf{R}^{\mathcal{C}_0}_{\mathcal{W},t}
\end{equation}
where $\mathbf{p}^{\mathcal{C}_0}_{\mathcal{W},t}$ and $\mathbf{R}^{\mathcal{C}_0}_{\mathcal{W},t}$ are the wrist translation and rotation at time $t$ in $\mathcal{C}_0$. We then assemble the wrist pose $\mathbf{T}_{\mathcal{R}\leftarrow\mathcal{W},t}\!\in\!\mathrm{SE}(3)$ and its quaternion parameterization $\mathbf{q}^{\mathcal{R}}_{\mathcal{W},t}$ for downstream control.

\textit{Robot Wrist and Finger Execution.}
With the wrist trajectory expressed in $\mathcal{R}$, we use cuRobo\cite{curobo_report23} to solve IK and obtain arm joint trajectories that follow $\{\mathbf{p}^{\mathcal{R}}_{\mathcal{W},t},\,\mathbf{R}^{\mathcal{R}}_{\mathcal{W},t}\}_{t=0}^{T-1}$. In parallel, the finger joint sequence $\{\mathbf{q}_{t}\}_{t=0}^{T-1}$ is sent directly to the robot hand controller. Finally, the robot control API executes these synchronized arm and hand trajectories to complete the task.

\section{Evaluating task-level generalization}



Traditionally, robot foundation models are evaluated on tasks similar to those in their training sets - picking up slightly different objects at randomized locations after seeing a lot of pick-and-place trajectories, or folding t-shirts after learning from large amout of t-shirt folding data~\cite{brohan2022rt, kim2024openvlaopensourcevisionlanguageactionmodel, pi0-experiment-wild}.  We refer to these as object-level and configuration-level generalizations. They are exciting steps towards a zero-shot model but the ``verbs'' in the task description are constrained to this small set, like ``pick'' or ``fold''.

We are interested in a stronger type of generalization - zero-shot task-level generalization: evaluating whether a model can perform drastically different tasks it has never encountered.



\subsection{Third-party selection of novel tasks}

We believe that true task-level generalization should allow any human to propose a task in any environment—without requiring prior knowledge of the capability of the model.
To this end, we crowdsource test data from third-party participants by asking them to propose manipulation tasks from their everyday surroundings. Each participant was instructed to:
(1) propose a short manipulation task that takes 3–5 seconds for a human;
(2) take a photo of the scene showing both the hand and the target object; 
(3) write a brief text description of the intended task; and
(4) stay diverse and challenging, be creative about tasks and scenes.

After this step, we gathered around 200 tasks with very out-of-distribution scene like ``at a gasoline pump'', out-of-distribution yet hard tasks such as ``flush the toilet'' or ``tear the tape''. However, we noticed that some volunteers still submitted low quality data such as blurry photos or boring tasks. To ensure quality, a separate group of third-party annotators filtered out samples that did not follow instructions or resembled basic tabletop pick-and-push tasks already covered in existing robot datasets. After filtering, 100 high-quality tasks remained, each consisting of one observation image and an instruction text. The instruction texts were further refined using Gemini to produce more detailed task descriptions.

\subsection{Evaluating video motion planning}
\begin{figure}[t]
    \centering
    \includegraphics[width=\linewidth]{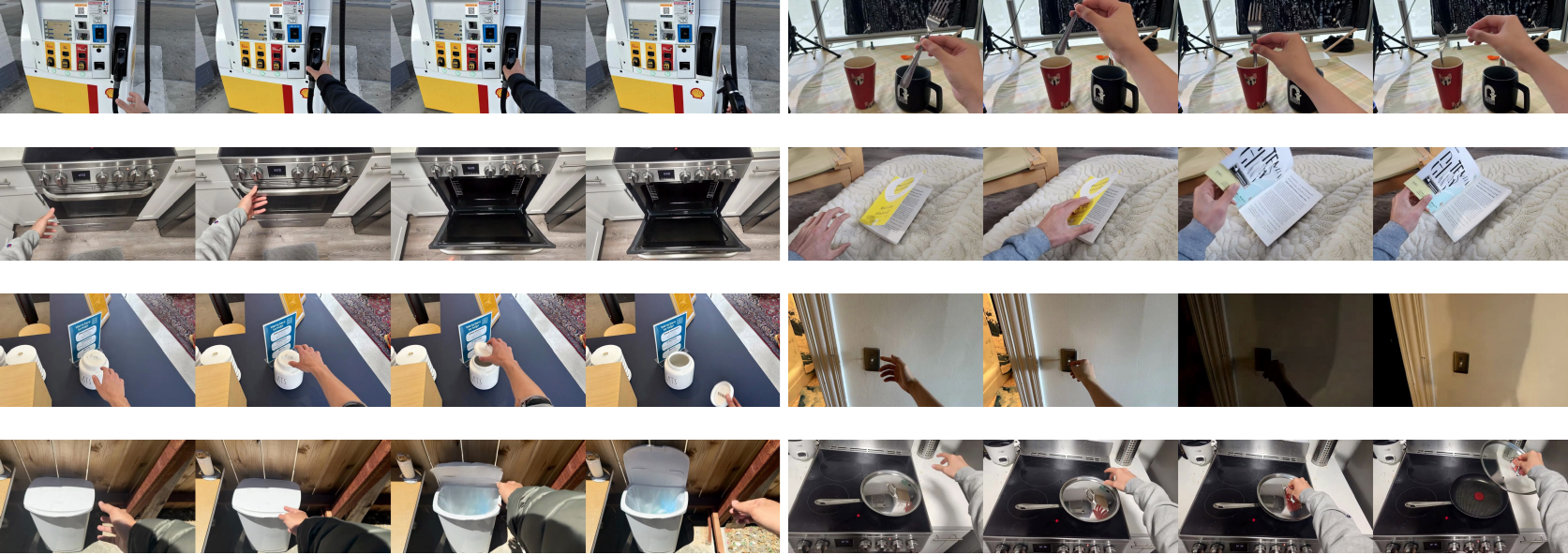}
    \put(-345, 105){\capfont  Grab the black gas nozzle}
    \put(-150, 105){\capfont  Place the fork into the red cup}
    \put(-345, 68){\capfont Open the oven}
    \put(-150, 68){\capfont Open the book cover}
    \put(-345, 31){\capfont  Lift the lid off the white jar}
    \put(-150, 31){\capfont  Switch off the light}
    \put(-345, -7){\capfont  Open the trash bin}
    \put(-150, -7){\capfont  Lift the pan lid}
    \vspace{-8pt}
    \caption{\textbf{Visualization of generated video plans. } Eight examples in our in-the-wild test set with generated videos by LVP. }
    \label{fig:eval_data}
\end{figure}

\begin{figure}[t]
    \centering
    \includegraphics[width=\linewidth]{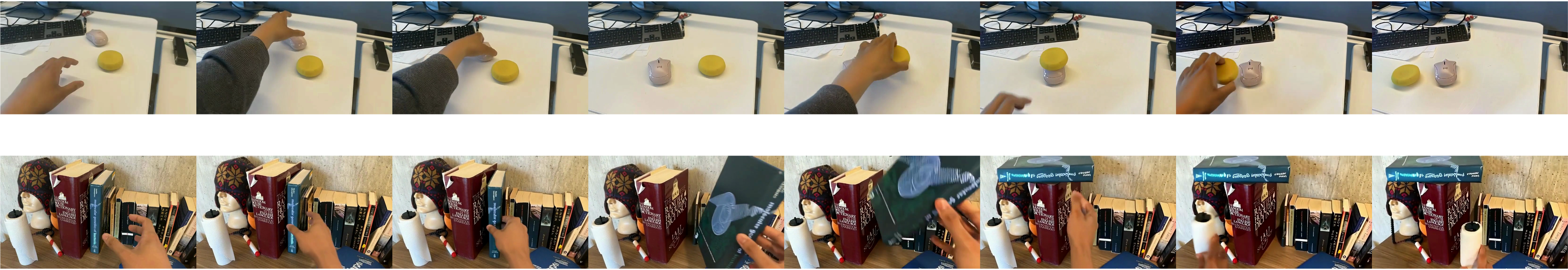}
    \put(-375, 31){\capfont Move the mouse closer}
    \put(-260, 31){\capfont  Stack the yellow object on the mouse}
    \put(-100, 31){\capfont  Move it on the table}
    \put(-375, -7){\capfont  Pull out the blue book}
    \put(-260, -7){\capfont  Stack it on top of the red book}
    \put(-100, -7){\capfont  Take the paper towel}
    \vspace{-3pt}
    \caption{\textbf{Multi-Stage Video Plans}.  Our LVP can generate long-horizon video plans by repeatedly extending videos conditioned on the last six latent frames. Each example illustrates a three-stage motion plan obtained through two iterative extensions. }
    \label{fig:video_multistage}
    \vspace{-5pt}
\end{figure}
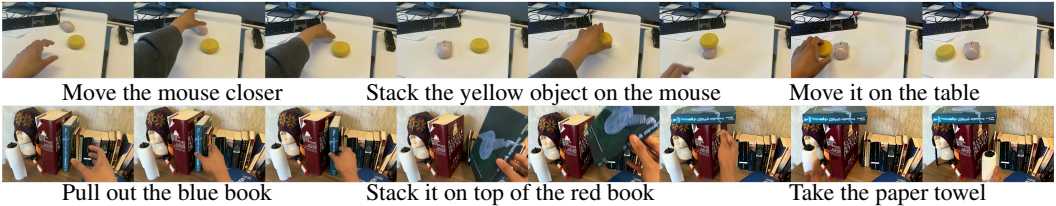

We first evaluate the stand-alone performance of our video planner. We feed oberservation images and rephrased instruction texts to our model. Figure~\ref{fig:eval_data} shows qualitative examples of generated video plans on this in-the-wild test set. Figure~\ref{fig:video_multistage} shows multi-stage video plans by extending generated videos repeatedly using our video-to-video generation.

\begin{figure}[t]
    \centering
    \includegraphics[width=\linewidth]{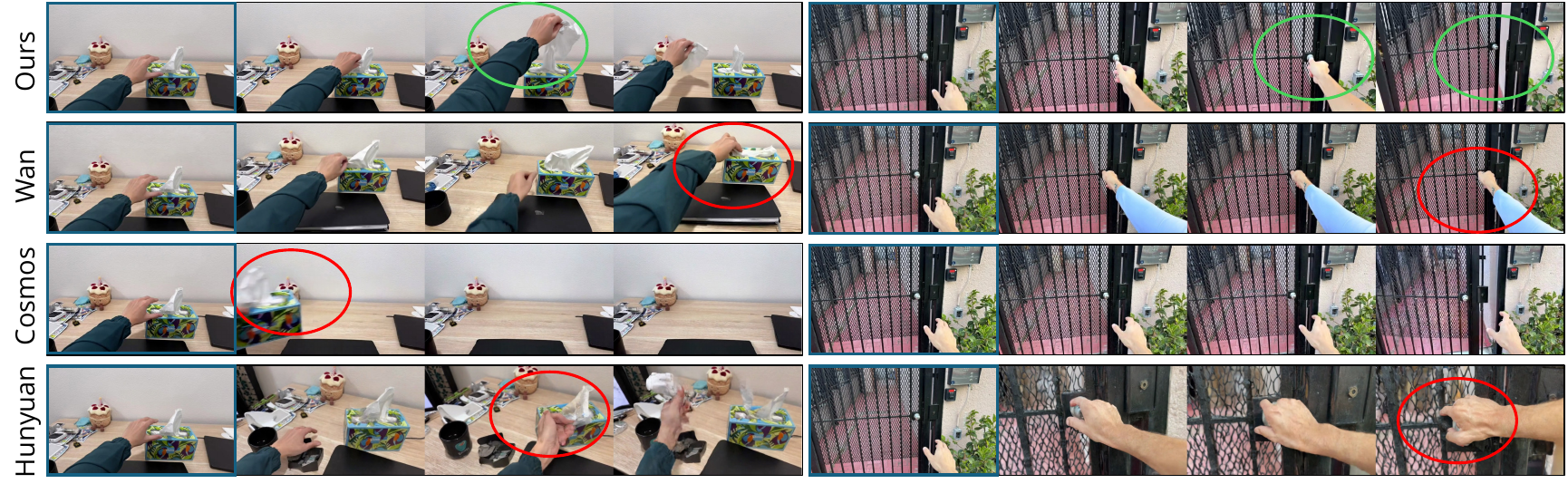}
    \vspace{-20pt}
    \put(-110, 15){\capfont \textbf{(a)} Pull out the tissue}
    \put(90, 15){\capfont \textbf{(b)} Open gate}
    \caption{\textbf{Baseline Comparison.} LVP accurately generates videos of hand interactions in a zero-shot setting, such as pulling out a tissue (left) and opening a gate (right). Baseline models (Wan, Cosmos-Predict 2, Hunyuan) often produce spatial or semantic inconsistencies, highlighted by red circles. The first frame and task instruction shown under each column serve as the generation conditions.  }
    \label{fig:video_results}
    \vspace{-5pt}
\end{figure}

We compare against three strong video generation baselines: Wan 2.1 I2V 14B~\citep{wan2025}, Cosmos-Predict 2 14B~\citep{agarwal2025cosmos}, and Hunyuan I2V 13B~\citep{kong2024hunyuanvideo}.   For each prompt, every method generates four videos.  

We design a four-level evaluation metric that measures instruction following, motion planning feasibility, and physical realism.

\begin{itemize}[leftmargin=2em, itemsep=0.05em]
     \item[{\bf (1)}] \textit{Correct contact}: The hand makes contact with the specified object at a correct location. Failures include touching the wrong object or making no contact. 
    \item[{\bf (2)}] \textit{Correct end state}: The final frame achieves the instructed goal (motion quality ignored).
    \item[{\bf (3)}] \textit{Task complete}:  Both correct contact and correct end state with plausible, continuous motion (minor physics artifacts allowed). 
    \item[{\bf (4)}] \textit{Perfect task complete}: The task is completed with visually flawless physics and no noticeable artifacts.  This highest level incorporates all prior criteria and additionally evaluates physical consistency and visual fidelity.
\end{itemize}
Levels 1–2 test comprehension and prompt following—whether the model correctly interprets and interacts with the right objects. Level 3 evaluates whether the model can generate complete video planning with feasible and coherent motions. Level 4 additionally measures physical realism and overall visual fidelity.

We ask third-party annotators to score all the generated videos and report both the average success rate and Best@4 (best result among four generations) for each level in Table~\ref{tab:video_eval}, with quantitative comparisons illustrated in Figure~\ref{fig:video_results}.

For all models, performance decreases monotonically from Level 1 to Level 4, reflecting the increasing difficulty of each criterion.
While pretrained Wan 2.1 achieves relatively high scores on Level 1 (correct contact), its performance drops sharply on Levels 2–4, indicating that it can initiate the correct interaction but fails to produce coherent, task-complete motion trajectories.
In contrast, our model achieves significantly higher scores across all levels, with the largest gains at Levels 3 and 4, indicating better generalization in producing coherent, physically consistent motion planning under in-the-wild conditions. Notably, our model attains a 59.3\% success rate at Level 3 (Task Complete) on the third-party test set, highlighting its ability to perform coherent and semantically grounded motion planning for unseen tasks in unseen environments. In addition, in Figure~\ref{fig:video_multistage}, we illustrate how our model is able to rollout long video plans.

\begin{table}[t]
\centering
\small
\resizebox{\linewidth}{!}{
\begin{tabular}{l *{8}{r}}
\toprule
\multirow{2}{*}{Method} 
& \multicolumn{2}{c}{Level 1: Correct contact} 
& \multicolumn{2}{c}{Level 2: End state} 
& \multicolumn{2}{c}{Level 3: Task complete} 
& \multicolumn{2}{c}{Level 4: Perfect} \\
\cmidrule(lr){2-3}\cmidrule(lr){4-5}\cmidrule(lr){6-7}\cmidrule(lr){8-9}
& Average (\%) & Best@4 & Average (\%) & Best@4 & Average (\%) & Best@4 & Average (\%) & Best@4 \\
\midrule
Wan 2.1 I2V 14B      
& 83.9 & 99.0 & 47.0 & 80.0 & 39.3 & 76.0 & 20.5 & 53.0 \\
Cosmos-Predict 2 14B                
& 45.3 & 81.0 & 11.9 & 35.0 &  7.5 & 24.0 &  2.5 &  9.0 \\
Hunyuan I2V 14B 
& 68.7 & 96.0 & 27.3 & 65.0 & 13.5 & 42.0 &  7.2 & 27.0 \\
\midrule
\textbf{Ours}                       
& \textbf{87.3} & \textbf{100.0} & \textbf{63.2} & \textbf{85.0} 
& \textbf{59.3} & \textbf{82.0} & \textbf{44.0} & \textbf{71.0} \\
\bottomrule
\end{tabular}
}
\vspace{-5pt}
\caption{\textbf{Video Plan Evaluation.} Evaluation on 100 in-the-wild manipulation prompts collected from third-party participants. We report the average success rate (Average) and Best@4 for each level. Our method achieves substantially higher success at Levels 3–4 than the baselines, indicating stronger generation of coherent, task-complete plans in in-the-wild settings. }
\label{tab:video_eval}
\end{table}

\subsection{Evaluating Real-World Robot Manipulation}

The previous experiment demonstrates that our large video planner exhibits strong zero-shot generalization for unseen tasks and novel scenes.
We now evaluate the complete pipeline, from video generation to action retargeting and execution, on real-world robotic platforms.

\begin{figure}[t!]
  \centering
  \begin{minipage}[t]{0.55\textwidth}
    \centering
    \resizebox{1\linewidth}{!}{
    \begin{tabular}{l|ccc}
\hline
\textbf{Task Set and Tasks}&\textbf{Ours}& $\mathbf{\pi_0}$ & \textbf{OpenVLA} \\
\hline
Task Group A: w/ Parallel Gripper \\
\hline
Pick Objects                     & \textbf{5}/10  & 3/10 & 0/10 \\
Pick A into B                    & \textbf{3}/10  & 1/10 & 0/10 \\
Open Drawer                      & \textbf{2}/10  & 1/10 & 0/10 \\
Press Button                     & \textbf{4}/10  & 0/10 & 0/10 \\
\hline
Task Group B: w/ Dexterous Hands \\
\hline
Pick Objects   & \textbf{4}/10 & N/A & N/A \\
Press Elevator Button & \textbf{4}/5 & N/A & N/A \\
Sweep Tennis Ball into Bucket & \textbf{5}/5 & N/A & N/A\\
Open Box   \textcolor{red}{(a)}     & \textbf{2}/10 & N/A & N/A \\
Open Door  \textcolor{cyan}{(b)}      & \textbf{6}/10 & N/A & N/A \\
Wipe Table \textcolor{green}{(c)}      & \textbf{8}/10 & N/A & N/A \\
Scoop Coffee Beans \textcolor{orange}{(b)}    & \textbf{3}/5 & N/A & N/A\\
Tear off Clear Tape \textcolor{purple}{(b)}    & \textbf{2}/5 & N/A & N/A\\
\hline
Task Group C: Out-of-distribution Set\\
\hline
Pick Objects (OOD Object$^1$)    & \textbf{4}/10  & 0/10 & 0/10 \\
Pick A into B (OOD Object$^1$)   & \textbf{2}/10  & 0/10 & 0/10 \\
Pick Objects (OOD Scene$^2$)     & \textbf{6}/10  & 1/10 & 0/10 \\
Pick A into B (OOD Scene$^2$)    & \textbf{1}/10  & 0/10 & 0/10 \\
\hline
\end{tabular}
}
    \label{tab:comparison}
  \end{minipage}
  \hfill
  \begin{minipage}[t]{0.43\textwidth}
  \vspace{-32mm}
    \centering
    \includegraphics[width=1.1\textwidth]{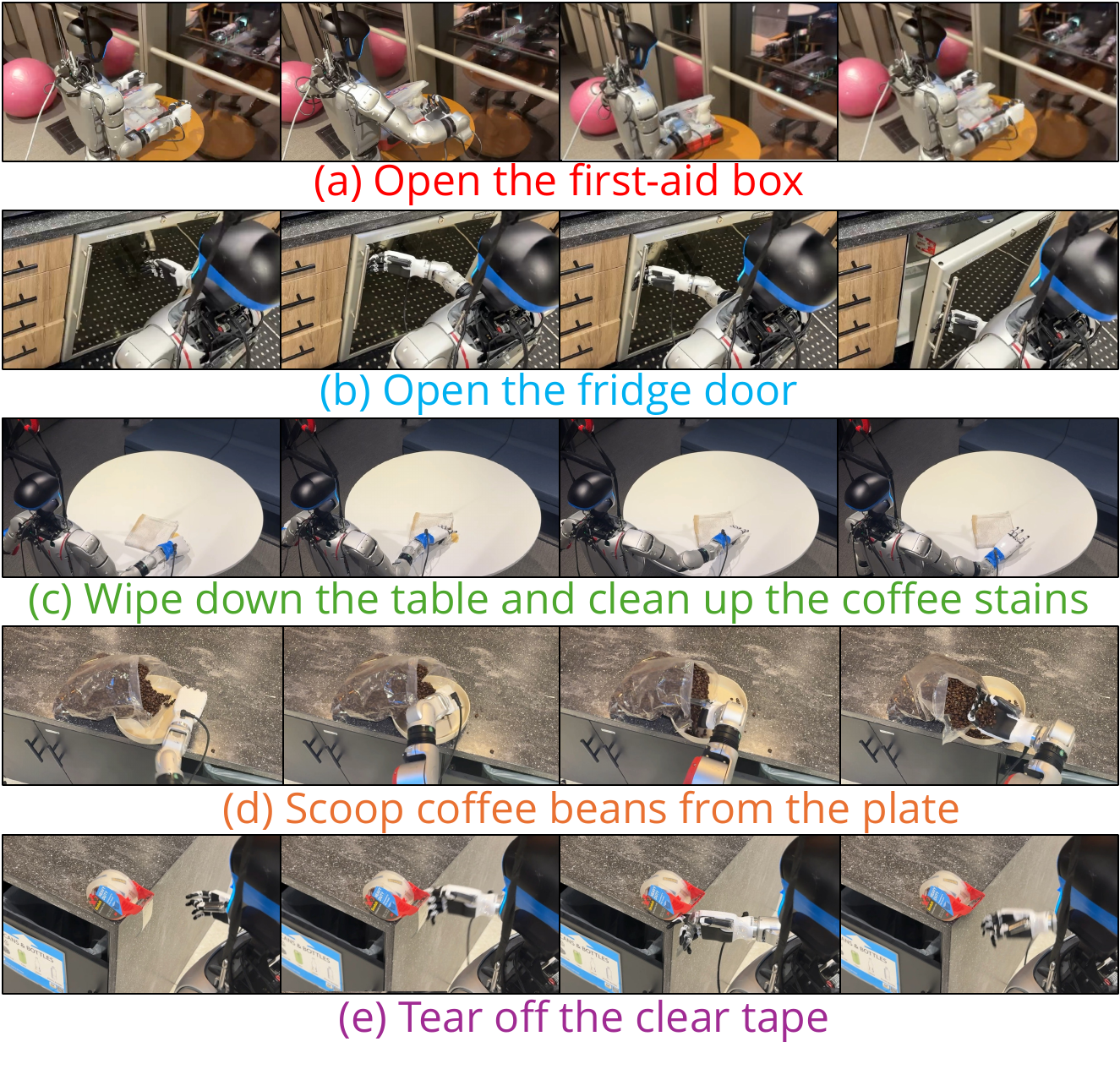}
  \end{minipage}
  \vspace{-10pt}
  \caption{\textbf{Robot Execution Evaluation.} Left: Comparison of Task Success Across Methods on (1) Franka Arm with Parallel-Jew Gripper and (2) G1 with Inspire Hands. $^{1}$ denotes tests on OOD objects; $^{2}$ denotes scenes that differ substantially from the training videos. Right: Visualization of the robot tasks and experiments. }
  \label{fig:robot_exp_example}
  \vspace{-15pt}
\end{figure}

\paragraph{Tasks}
We conduct experiments on two distinct robot morphologies: a Franka Emika Arm with a parallel-jaw gripper and a G1 Arm equipped with an Inspire dexterous hand.
Each platform is tested on task sets that highlight different manipulation capabilities.
For the dexterous hand, we further evaluate challenging novel tasks such as opening a door, opening a box, and scooping coffee beans, as shown in the right columns of Figure~\ref{fig:robot_exp_example}.

\begin{figure}[t]
    \centering
    \includegraphics[width=\linewidth]{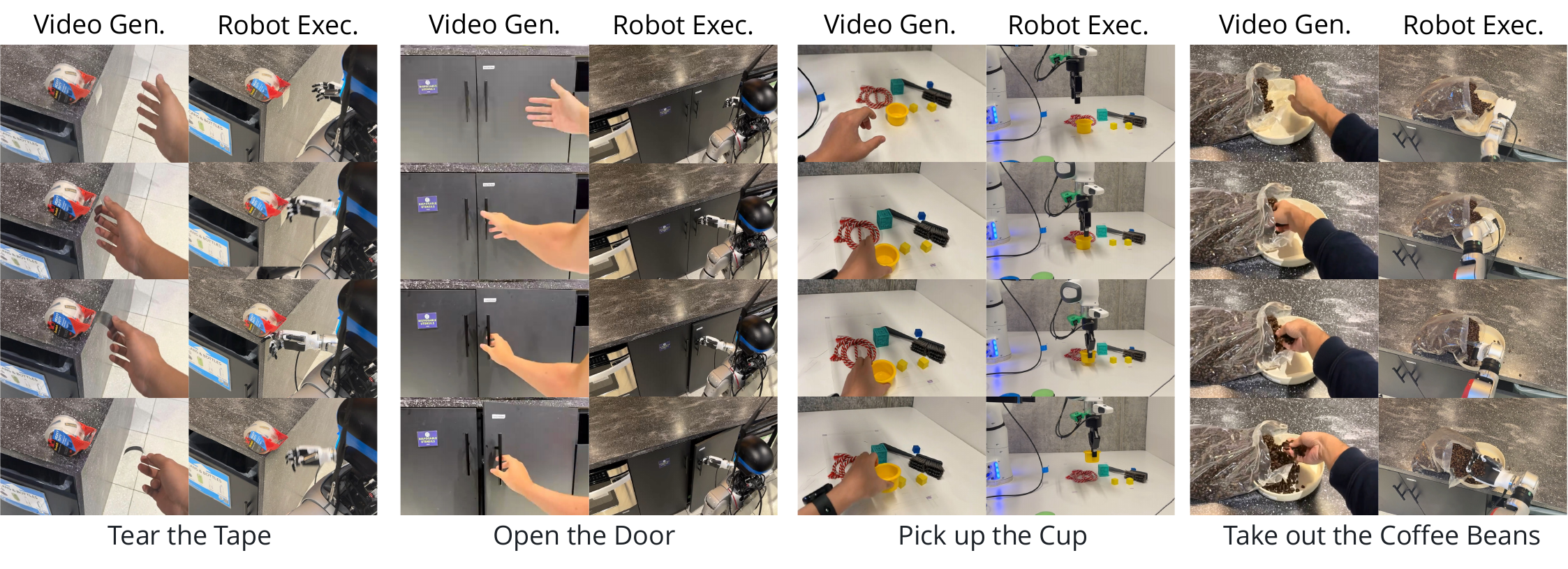}
    \vspace{-20pt}
    \caption{\textbf{Zero Shot Robot Manipulation with LVP.} The model generates videos for diverse tasks, enabling zero-shot execution on both a dexterous hand and a parallel-jaw gripper. }
    \label{fig:video_paired_robot}
\end{figure}


\begin{itemize}[leftmargin=2em, nosep]
\item \textit{Franka Arm with Parallel-Jaw Gripper }  As reported in Task Groups A and C of Table~\ref{fig:robot_exp_example},
this set focuses on manipulation tasks that can be achieved with simple two-finger grasps, such as object pick-and-place, block stacking, and bottle relocation. These tasks emphasize grasp detection and robust trajectory execution under limited actuation. Group A serves as the standard benchmark, while Group C primarily focuses on out-of-distribution task sets that contain unseen scenarios in the training video dataset. We include more details about gripper cases in Appendix~\ref{appendix: gripper}.

\item \textit{Humanoid with Dexterous Hand }  
As reported in Task Group B of Table~\ref{fig:robot_exp_example}
, this set targets fine-grained dexterous manipulation requiring multiple degrees of freedom. Tasks include in-hand rotation, tool use (e.g., pen writing or screwdriver insertion), and precise placement of irregular objects. These tasks stress the ability of our method to transfer complex human hand articulations to the robot hand.
\end{itemize}

\paragraph{Baselines}
We compare our method against several state-of-the-art vision-language-action baselines:  
\begin{itemize}[leftmargin=2em, nosep]
\item \textit{$\pi_0$}~\citep{black2024pi_0}:  
We evaluate $\pi_0$ model by loading the released checkpoint and directly testing its generalization to our benchmark tasks, following the standard usage protocol.
\item \textit{OpenVLA}~\citep{kim2024openvlaopensourcevisionlanguageactionmodel}:  
We include OpenVLA with its released checkpoint as a baseline, evaluating its performance on our task sets without additional fine-tuning.
\end{itemize}
Note that $\pi_0$ and OpenVLA are not compatible with multi-finger dexterous hand settings and are therefore only tested on parallel-gripper tasks.

\paragraph{Results}
Quantitative comparisons are presented in Table~\ref{fig:robot_exp_example}, and qualitative examples of successful executions are shown in Figure~\ref{fig:video_paired_robot}. Our approach exhibits strong zero-shot generalization on the most challenging settings—e.g., scooping coffee beans and tearing tape—underscoring the significance of the proposed method. Across both task suites, it consistently outperforms existing baselines, with especially outstanding performance on dexterous manipulation.
In contrast, baselines show strong performance at tasks similar to training distributions, e.g. picking up objects,  but struggle with task-level generalization. We speculate that such regression arises because imitation learning based robot foundation models have seen a lot of trajectories of pick-and-place, but have never seen enough diverse tasks to robustly generalize to new ones.

\section{Limitations} 
Our approach has several limitations. On the video generation side, producing a single video plan takes several minutes on a single A100 GPU, making direct real-time deployment on robots intractable. Potential solutions include step-distillation methods~\cite{salimans2022progressive, yin2024one}, which reduce the number of inference steps, or causal video models~\cite{chen2024diffusion, yin2025slow, huang2025self}, which lower the latency of the generation process. In addition, on the robotics side, our robot action extraction has several limitations. Our current robotics action extraction pipeline uses open-source models to estimate 4D reconstructions and hand pose estimations. Both of these models can make mistakes, sometimes leading to task failures. Even if all models succeed, the retarget might not be sufficient for certain dexterous hands. In addition, retargeting actions to the parallel-jaw gripper can be challenging due to its much lower degree-of-freedom count compared to a human hand.  Finally, our overall robot execution framework is run in an open-loop manner, which is not sufficient for accomplishing dexterous tasks.




\section{Conclusion} We investigate a different approach to robot foundation models with video as the backbone. We present Large Video Planner (LVP), a 14-billion parameter video foundation model for embodiment planning.
LVP generates videos as motion plans conditioned on one or a few scene frames and a text description of the task.
We demonstrate that these generated motion plans can be successfully retargeted to dexterous robotic hands using open-source reconstruction and retargeting tools.
Evaluations on third-party proposed tasks show evidence of task-level generalization, a capability limited in existing VLA models. We open-source our model, data, and training code to support the research community and hope this work will inspire further exploration of video foundation model for robotics.

\clearpage
\bibliography{main}
\bibliographystyle{iclr2026_conference}
\clearpage
\clearpage
\appendix
\section{Details of \dataname Dataset}

\label{app:data_details}

\subsection{Extracting hand interaction videos from Pandas}
\label{app:subsec_pandas}

As summarized in Table~\ref{tab:dataset_collections}, we extract a subset of 196K video clips focusing on human hand interactions from the Pandas-70M dataset~\citep{chen2024panda}. The process is as follows.

We begin with the metadata from Pandas-70M, which provides short text captions for each clip. Using a whitelist of 108 keywords and a blacklist of 84 keywords, we perform an initial keyword-based filtering. After this step, we download approximately 692K videos of variable length. 

Each video is segmented into multiple non-overlapping 4-second clips, and we run human pose detection at 1 FPS on frames resized to $768\times1024$. We retain only clips containing 1–3 valid human detections, where a detection is considered valid if the bounding box height is $\leq$ 60 pixels and width is $\geq$ 120 pixels.

Next, we use Gemini-2.0 Flash to caption and evaluate each video based on four questions:

\begin{itemize}
    \item  (1) Does the video contain rich human hand motions?
    \item (2)
Does the video show a human performing any meaningful actions (e.g., manipulating objects, using tools, cooking,  pushing or pulling objects, waving, clapping, washing hands, etc.)? 
   \item (3) Is the video playing at a normal speed? 
   \item (4) Does the video contain scene changes (i.e., camera shot changes or viewpoint changes)? 
\end{itemize} 
We retain only videos with answers True, True, True, and False to these four questions. Finally, we retain 196K clips. 

\subsection{Data preprocessing pipeline}

\begin{figure}[h]
    \centering
    \includegraphics[width=\linewidth]{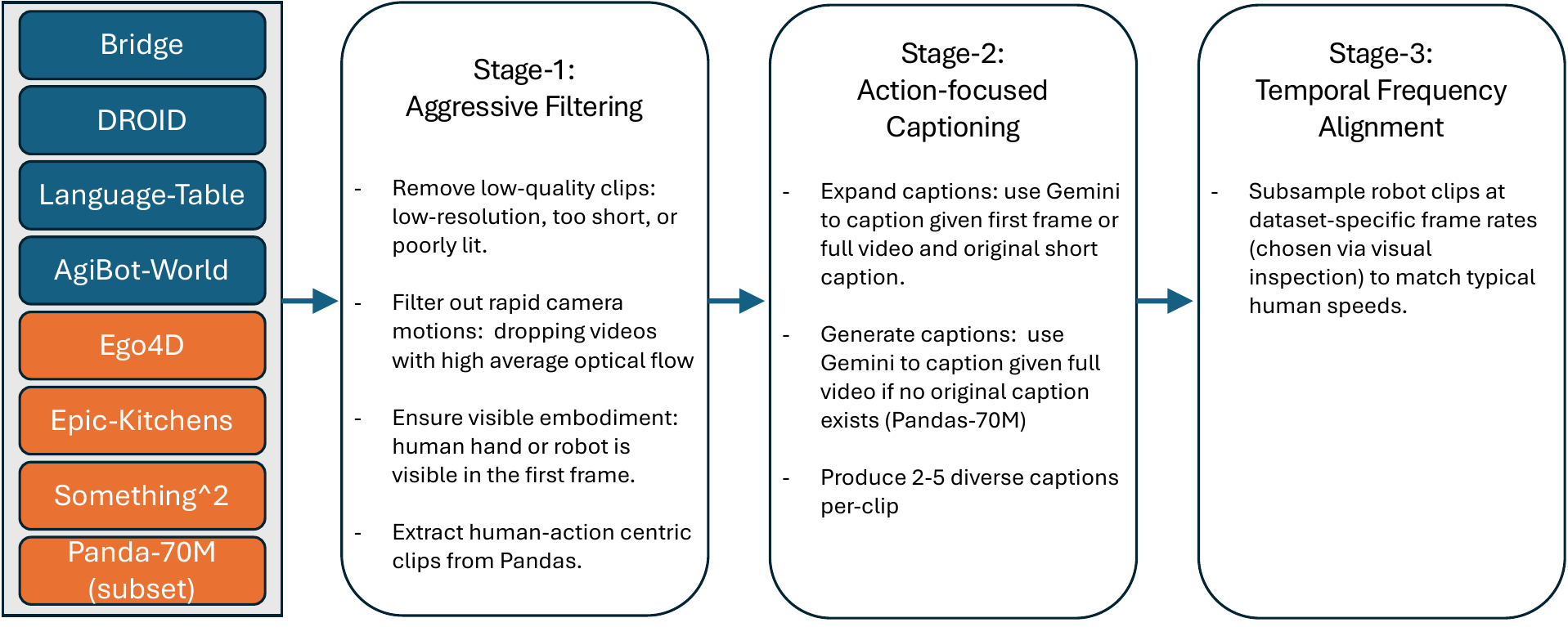}
    \vspace{-8pt}
    \caption{\textbf{\dataname curation pipelines.} Videos are collected from eight public sources, including four teleoperated robotics datasets and four human-centric activity datasets. 
    We apply three processing stages: (1) aggressive filtering for quality and embodiment, (2) action-focused captioning using Gemini, and (3) temporal frequency alignment to match human motion speeds. 
    The final dataset contains 1.4 million clips with diverse, action-centric text captions. }
    \label{fig:appendix_data_pipeline}
\end{figure}

\begin{figure}[t]
    \centering
    \includegraphics[width=0.8\linewidth]{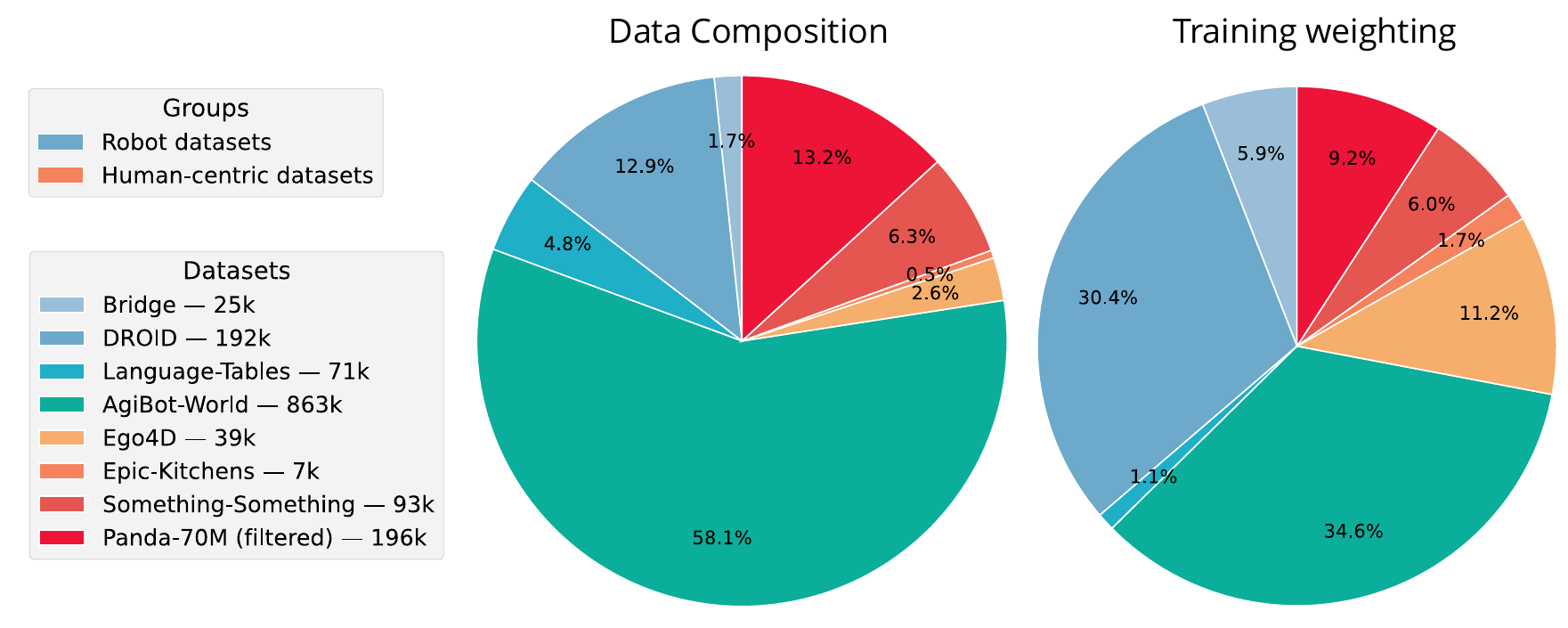}
    \vspace{-8pt}
    \caption{\textbf{Composition of \dataname.}
Distribution of filtered \dataname clips across eight sources.
Greenish tones indicate teleoperated robot datasets, while reddish tones represent human-centric activity datasets. The rightmost column shows the final sampling ratios after reweighting during first-stage training. }
    \label{fig:appendix_data_pie}
\end{figure}

Here, we illustrate the data curation pipleine in Figure~\ref{fig:appendix_data_pipeline}. We also illusrate the final composition of the \dataname using a pie-chart in Figure~\ref{fig:appendix_data_pie}. We also include some details about each stage below.

\paragraph{Filtering our rapid camera motions} Lot's of egocentric videos from Ego4D, Epic-Kitchens contains large camera motions,  this hinders the model from learning forgound actions and we hope to remove majority of them. To do so,  we compute per-pixel optical flow using OpenCV\footnote{\url{https://opencv.org/}} at 4 frames per second on resized frames of $256 \times 256$ resolution, and remove the top 30\% of videos with the highest spatio-temporal average flow magnitude.

\section{Details of Video Generative Model Training}

Each training sample consists of a 49-frame video clip at a resolution of $832\times480$. The clip is encoded into VAE latent at shape of $104\times60\times13$. In the continue pretraining stage, we train the model with batch size of 128 for 60,000 iterations, we use a constant learning rate of $1 \times 10^{-5}$ after 1000 steps of warmup. During the finetuning stage, we use the same batch size and trains for 10,000 iterations with a reduced learning rate at $2.5 \times 10^{-6}$.

\paragraph{Data reweighting} As shown in Table~\ref{tab:dataset_collections}, the eight dataset sources vary greatly in size. To balance their contributions during training, we employ weighted sampling with respective weights of $0.375$, $0.75$, $1.5$, $0.5$, $0.5$, $1.0$, $2.0$, and $0.05$ for AgiBot-World, DROID, Ego4D, Pandas (filtered), SomethingSomething, Bridge, Epic-Kitchens, and Language Table. The resulting sampling composition is visualized in Figure~\ref{fig:appendix_data_pie}.

\section{Method: Comparison between HaMeR and Our 4D Alignment Module}

\begin{figure}[h]
    \centering
    \includegraphics[width=0.8\linewidth]{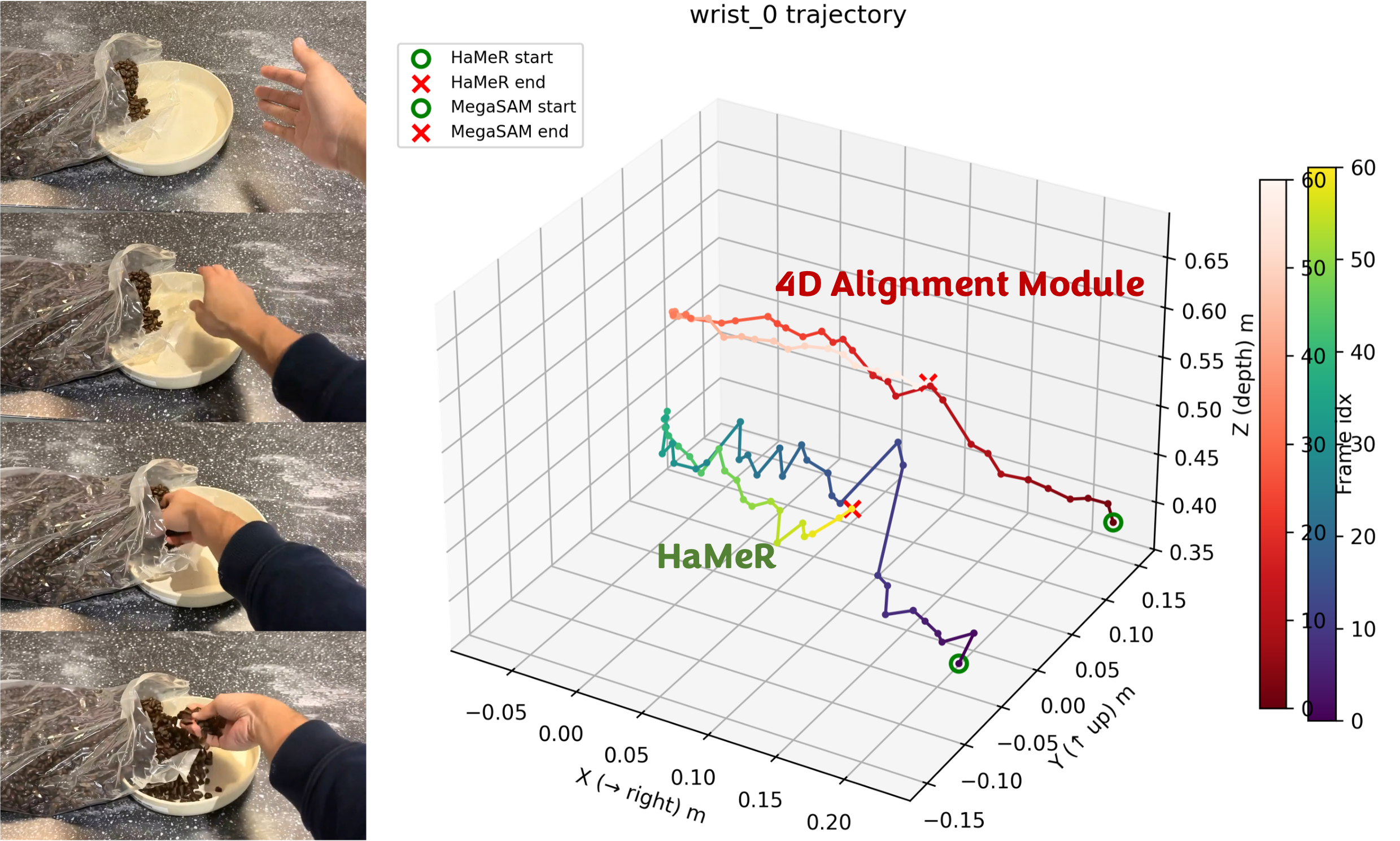}
    \caption{\textbf{Comparison between HaMeR and Our 4D Alignment Module.} 
    The red curve shows our 4D Alignment Module; the blue one is from HaMeR.}
    \label{fig:comparison_supp}
\end{figure}

In HaMeR, the depth estimation module lacks accuracy in metric scale estimation. Also, the hand hand pose is estimated with no temporal consistency. Hence, introducing MegaSaM would improve accuracy in depth estimation and smoothness in temporal hand trajectory estimation. To evaluate the effectiveness of our 4D Alignment module, we conduct an ablation study highlighting its ability to enforce spatio-temporal consistency, see Fig.~\ref{fig:comparison_supp}.

\section{Method: Details about Our Video Processing Modules}
\subsection{Data Acquisition and Input}
Given a monocular RGB sequence $\{I_t\}_{t=1}^{T}$ without known camera intrinsics or extrinsics, our objective is to recover a metrically consistent, temporally smooth hand motion representation that includes (i) per–frame joints angles and (ii) a wrist pose trajectory in a fixed world frame.

\subsection{HaMeR: Hand Mesh and Camera Estimation}
For each frame, HaMeR~\citep{hamer} predicts MANO parameters, hand mesh vertices, and weak-perspective camera parameters. These outputs provide articulated 3D joint positions and their 2D projections, together with a global wrist orientation and translation, forming the per-frame hand motion estimates. 
However, due to the inherent scale ambiguity of monocular weak-perspective modeling and the lack of temporal constraints, HaMeR’s predicted translations are often noisy and drift over time. 
As a result, the recovered wrist trajectories are not metrically accurate and cannot be directly used for real-world robot execution. 
To resolve these issues, we incorporate depth and multi-view cues from MegaSaM, which performs global optimization across frames to refine both the metric scale and temporal consistency of the estimated hand motion.

\subsection{MegaSaM: Depth-Assisted Structure Recovery}
To improve temporal consistency and resolve monocular scale ambiguity, we integrate MegaSaM~\citep{li2024megasamaccuratefastrobust}. Depth maps are estimated per frame and refined with bundle adjustment across the sequence, yielding per-frame extrinsics and metrically aligned depth. The first frame is chosen as the world reference system.

\subsection{Alignment of HaMeR and MegaSaM}
We first use MegaSaM to estimate the intrinsic parameters ${\mathbf{K}}$, which are then provided to HaMeR for more reliable per-frame wrist localization. 
Specifically, HaMeR outputs the wrist pixel coordinate $(u_t, v_t)$ in the image plane. 
We then sample the corresponding depth $z_t = D_t(u_t,v_t)$ from the MegaSaM depth map and back-project it using ${\mathbf{K}}$:
\begin{equation}
\mathbf{p}^{\text{cam}}_{t,w} = z_t\,{\mathbf{K}}^{-1}
\begin{bmatrix}u_t\\ v_t\\ 1\end{bmatrix}.
\end{equation}
The wrist position in the first frame $\mathbf{p}^{\text{w}}_{t,w}$ is recovered by applying the optimized extrinsics $\mathbf{T}_t=[\mathbf{R}_t|\mathbf{t}_t]$ from MegaSaM:
\begin{equation}
\mathbf{p}^{\text{w}}_{t,w} = \mathbf{R}_t\,\mathbf{p}^{\text{cam}}_{t,w} + \mathbf{t}_t.
\end{equation}

For Camera-to-Robot Alignment, we note the first camera frame $\mathcal{C}_0$, the MANO frame $\mathcal{M}$, and the robot frame $\mathcal{R}$.
We apply the rotation $\mathbf{M}\in\mathrm{SO}(3)$ to unify axes between $\mathcal{C}_0$ and $\mathcal{R}$. Translations use the same linear map:
\begin{equation}
\label{eq:frame-map}
\mathbf{p}^{\mathcal{R}}_{\mathcal{W},t}=\mathbf{M}\,\mathbf{p}^{\mathcal{C}_0}_{\mathcal{W},t},\qquad
\mathbf{R}^{\mathcal{R}}_{\mathcal{W},t}=\mathbf{M}\,\mathbf{R}^{\mathcal{C}_0}_{\mathcal{W},t}\,\mathbf{M}^\top.
\end{equation}
We then form the wrist pose $\mathbf{T}_{\mathcal{R}\leftarrow\mathcal{W},t}=\begin{bmatrix}
\mathbf{R}^{\mathcal{R}}_{\mathcal{W},t} & \mathbf{p}^{\mathcal{R}}_{\mathcal{W},t}\\
\mathbf{0}^\top & 1
\end{bmatrix}$ and its quaternion $\mathbf{q}^{\mathcal{R}}_{\mathcal{W},t}$.

\subsection{Wrist Pose and Robot Frame Transformation}
For execution on a real robot, the recovered wrist trajectories must be expressed in the G1 base coordinate system rather than the camera frame. 
To achieve this, we first estimate the extrinsic transformation $\mathbf{T}_{\text{G1}\leftarrow {w}}$, parameterized as a rotation $\mathbf{R}_{\text{G1}\leftarrow{w}} \in \mathrm{SO}(3)$ and a translation $\mathbf{t}_{\text{G1}\leftarrow{w}} \in \mathbb{R}^3$, between the reference camera world frame (camera in the first frame) and the G1 world frame, based on the known orientation of the G1 base and the viewing direction of the captured video. 
Applying this transformation maps each wrist position and orientation from the reference camera frame into the G1 base frame:
\begin{equation}
\mathbf{p}^{\text{G1}}_{t,w} = \mathbf{R}_{\text{G1}\leftarrow{w}}\,\mathbf{p}^{\text{w}}_{t,w} + \mathbf{t}_{\text{G1}\leftarrow{w}}
\end{equation}

For Temporal Completion and Smoothing, we give the formulation as:
\begin{equation}
\hat{\mathbf{p}}^{\mathcal{R}}_{\mathcal{W},t}=\mathrm{SG}\!\big(\tilde{\mathbf{p}}^{\mathcal{R}}_{\mathcal{W},:}\big)_t,\qquad
\hat{\mathbf{q}}^{\mathcal{R}}_{\mathcal{W},t}=
\frac{\mathrm{SG}\!\big(\mathbf{q}^{\mathcal{R}}_{\mathcal{W},:}\big)_t}{\left\lVert \mathrm{SG}\!\big(\mathbf{q}^{\mathcal{R}}_{\mathcal{W},:}\big)_t\right\rVert_2}.
\end{equation}
The final wrist pose is $\hat{\mathbf{T}}_{\mathcal{R}\leftarrow\mathcal{W},t}=\big(\hat{\mathbf{R}}(\hat{\mathbf{q}}^{\mathcal{R}}_{\mathcal{W},t}),\,\hat{\mathbf{p}}^{\mathcal{R}}_{\mathcal{W},t}\big)$.

\subsection{Other End-effectors Retargeting: Parallel-Jaw Grippers.}
\label{appendix: gripper}
We have already included how to generate five-finger robot hand motion in the main paper. Here we introduce how to generate motion for parallel-jaw grippers with our method. 
Retargeting a five-finger human hand directly to a two-finger gripper is inherently under-constrained, particularly for grasping motions.
To address this, we employ \textit{GraspNet}~\cite{fang2020graspnetlargescaleclustereddensely}
to predict candidate grasp poses consistent with the observed scene geometry.
Grasp execution is then triggered by heuristics that detect grasping intent from
the human hand motion, ensuring reliable transfer of manipulation behaviors
despite the reduced degrees of freedom.

\section{Experiments: Task Set and Tasks Description}

\begin{table}[H]
\centering
\small
\resizebox{\linewidth}{!}{
\begin{tabular}{l|c|p{8cm}}
\toprule
\textbf{Task Name} & \textbf{Robot Setup} & \textbf{Task Description} \\
\midrule
Press Button & (1) Franka+Gripper & A robot arm is required to locate and press a button initialized at any arbitrary position within its reachable workspace. \\
\hline
Pick Objects & (1) Franka+Gripper & A robot arm is required to locate, grasp, and lift objects placed at arbitrary positions within its reachable workspace for manipulation. \\
\hline
Pick A into B & (1) Franka+Gripper & A robot arm is required to grasp object A and accurately place it into container B located at arbitrary positions in workspace. \\
\hline
Open Drawer & (1) Franka+Gripper & A robot arm is required to grasp the handle and pull open a drawer to a designated position. \\
\hline
Pick Objects (OOD Object) & (1) Franka+Gripper & A robot arm is required to pick unfamiliar, out-of-distribution objects randomly placed within its workspace. \\
\hline
Pick A into B (OOD Object) & (1) Franka+Gripper & A robot arm is required to place novel, out-of-distribution objects into container B within the workspace. \\
\hline
Pick Objects (OOD Scene) & (1) Franka+Gripper & A robot arm is required to pick objects in unseen, out-of-distribution scenes with new layouts or obstacles. \\
\hline
Pick A into B (OOD Scene) & (1) Franka+Gripper & A robot arm is required to place objects into container B under unseen, out-of-distribution scene configurations. \\
\hline
\hline
Pick Objects & (2) G1+DexHand & A dexterous robot hand is required to locate, grasp, and lift objects from arbitrary positions within workspace. \\
\hline
Open Box & (2) G1+DexHand & A dexterous robot hand is required to grasp and open a box lid placed in the workspace. \\
\hline
Open Door & (2) G1+DexHand & A dexterous robot hand is required to grasp and pull open a door handle to a designated position. \\
\hline
Wipe Table & (2) G1+DexHand & A dexterous robot hand is required to wipe the table surface clean using an object or cloth. \\
\hline
Press Elevator Button & (2) G1+DexHand & A dexterous robot hand is required to press an elevator button at any arbitrary position in the workspace. \\
\hline
Scoop Coffee Beans & (2) G1+DexHand & A dexterous robot hand is required to use a scoop to collect coffee beans and lift them successfully. \\
\hline
Sweep Tennis Ball into Bucket & (2) G1+DexHand & A dexterous robot hand is required to sweep a tennis ball across the surface into a target bucket. \\
\hline
Tear off Clear Tape & (2) G1+DexHand & A dexterous robot hand is required to peel and tear off a strip of transparent adhesive tape. \\
\bottomrule
\end{tabular}
}
\vspace{-5pt}
\caption{Task Set and Tasks Description}
\label{tab:tasks}
\end{table}

\section{Experiments: Real-World Robots Set Up}

\begin{table}[H]
\centering
\small
\resizebox{0.8\linewidth}{!}{
\begin{tabular}{lcc}
\toprule
\textbf{Tasks} & \textbf{Robot} & \textbf{Control Frequency} \\
\midrule
Task Set1 & Franka Panda Arm + Parallel-Jaw Gripper & 15 Hz \\
Task Set2 & Unitree G1 Arm + Inspire Hand (DH56DFX) & 5 Hz \\
\bottomrule
\end{tabular}
}
\caption{Real-world robot experimental settings.}
\label{tab:realworld}
\end{table}

In Task Set 1, we used a Franka Emika Panda arm equipped with a parallel-jaw gripper for baseline grasping tasks.  

In Task Set 2, we combined a Unitree G1 Humanoid with an Inspire dexterous hand. The two were mechanically connected via a flange, and we implemented synchronized arm-hand control at 5~Hz based on the Unitree teleoperation\footnote{\url{https://github.com/unitreerobotics/xr_teleoperate}} framework. Furthermore, the joint angles predicted by our dex-retargeting module were remapped into the valid motor command ranges of the Inspire Hand to enable real-time execution.

When conducting the humanoid experiments, we used tape on the gripper in some trials to compensate for limited gripper torque or insufficient friction. We consider this type of gripper modification to be a common hardware adjustment, and it does not affect the main conclusions of the experiments.

\section{Video Results}

We provide additional qualitative video results. For each video, eight frames are uniformly sampled for visualization. The first frame represents the input observation image, and the text caption below each sequence shows the task instruction (prior to rephrasing by the language model).

\begin{figure}[h]
    \centering
    \includegraphics[width=\linewidth]{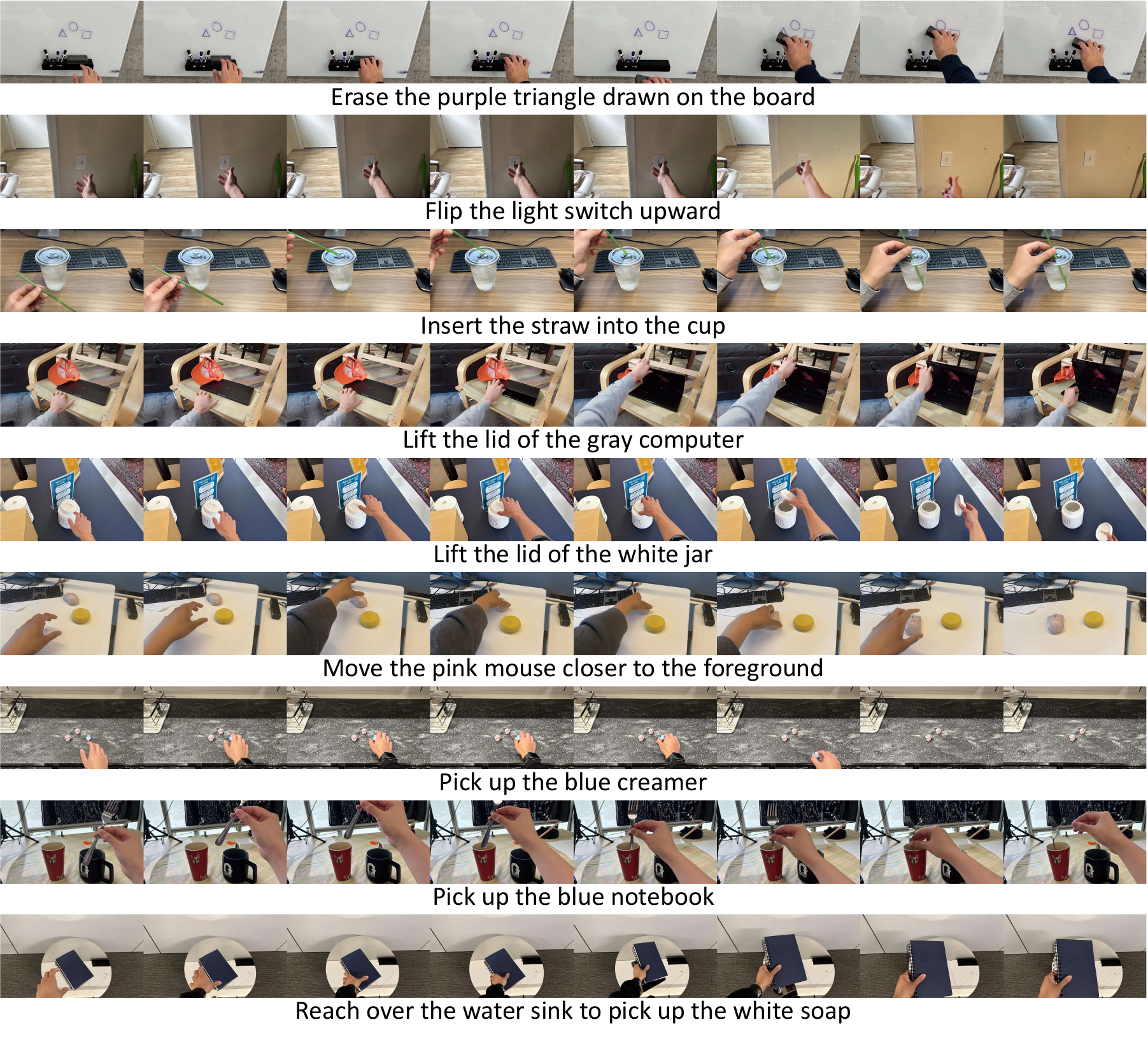}
    \vspace{-10pt}
    \caption{\textbf{Visualization of generated video plans with human hand. } Each row shows eight uniformly sampled frames from a generated video plan. The first frame depicts the input condition image that defines the scene. The caption below each sequence indicates the task instruction used to generate the video.}
    \label{fig:figure_supp_results_gallery_A}
\end{figure}

\begin{figure}[t]
    \centering
    \includegraphics[width=\linewidth]{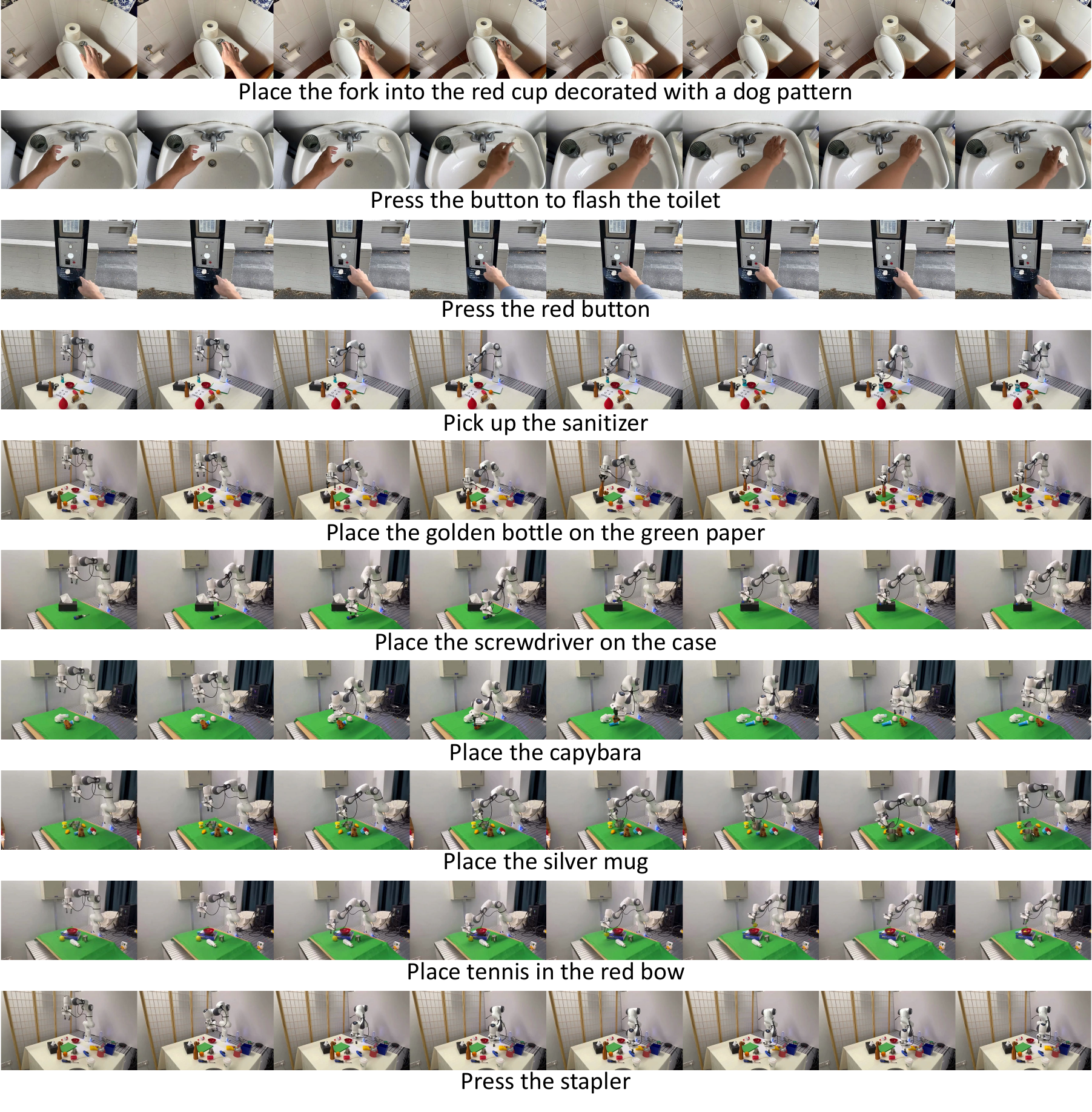}
    \vspace{-10pt}
    \caption{\textbf{Visualization of generated video plans with human hand and robot gripper. } Each row shows eight uniformly sampled frames from a generated video plan. The first frame depicts the input condition image that defines the scene. The caption below each sequence indicates the task instruction used to generate the video.}
    \label{fig:figure_supp_results_gallery_B}
\end{figure}

\begin{figure}[t]
    \centering
    \includegraphics[width=\linewidth]{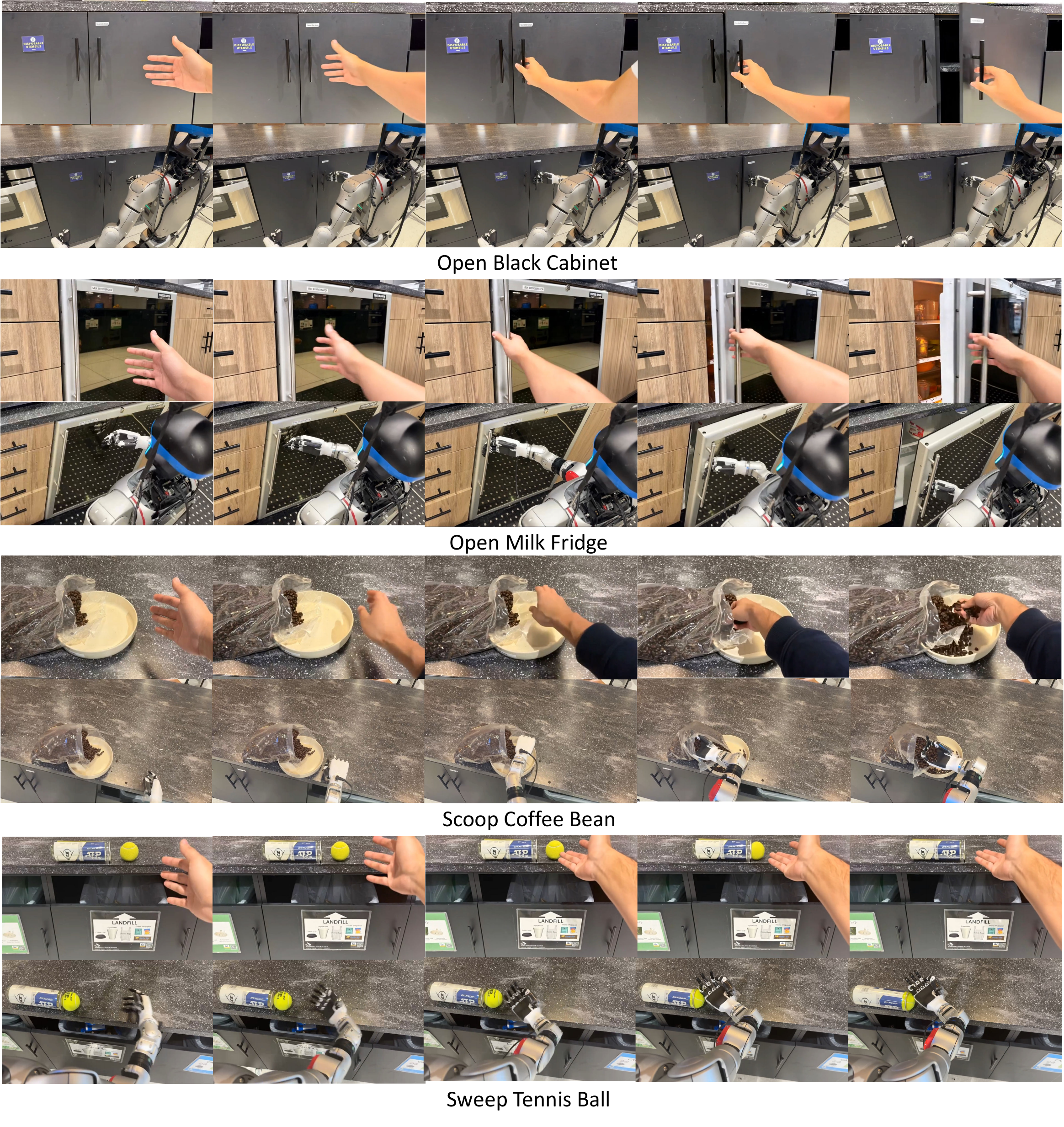}
    \vspace{-10pt}
    \caption{\textbf{Visualization of generated videos and corresponding robot executions -- A.} The first row presents five uniformly sampled frames from the generated video plan conditioned on the input scene image. The second row illustrates the robot executing the same task in real world.}
    \label{fig:figure_supp_results_gallery_B}
\end{figure}

\begin{figure}[t]
    \centering
    \includegraphics[width=\linewidth]{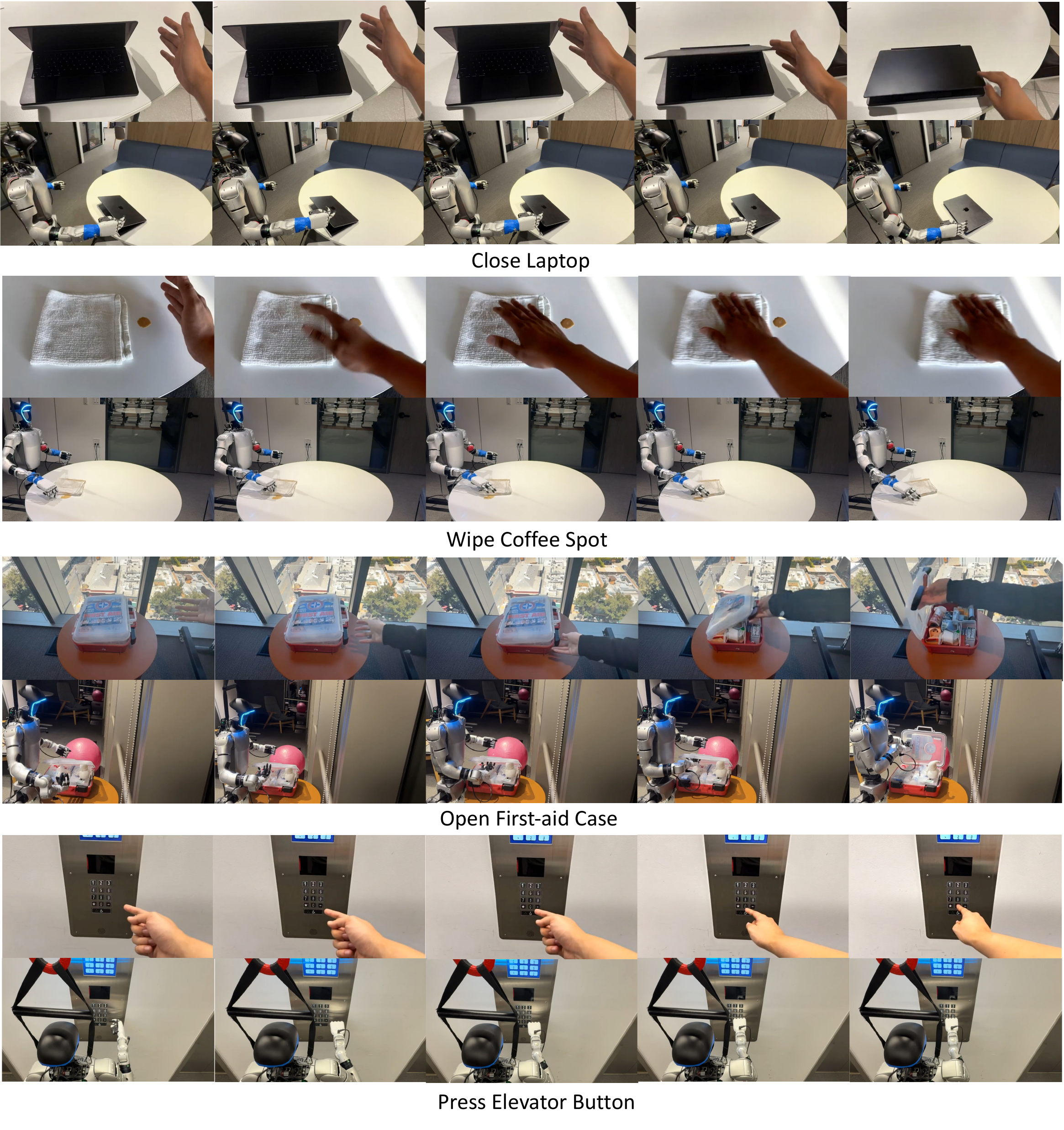}
    \vspace{-10pt}
    \caption{\textbf{Visualization of generated videos and corresponding robot executions -- B.} The first row presents five uniformly sampled frames from the generated video plan conditioned on the input scene image. The second row illustrates the robot executing the same task in real world.}
    \label{fig:figure_supp_results_gallery_B}
\end{figure}
\end{document}